\documentclass[lettersize,journal]{IEEEtran}
\usepackage{amsmath,amsfonts}
\usepackage{algorithmic}
\usepackage{algorithm}
\usepackage{array}
\usepackage{textcomp}
\usepackage{stfloats}
\usepackage{url}
\usepackage{verbatim}
\usepackage{graphicx}
\usepackage{cite}
\usepackage{tikz}
\usepackage{comment}
\usepackage{color}
\usepackage{subfigure}
\usepackage{amssymb}
\usepackage{booktabs}
\usepackage{mathrsfs}
\usepackage{multirow}
\usepackage{caption}
\usepackage{color}
\usepackage{colortbl}  
\usepackage{xcolor}
\usepackage{array}   
\usepackage[pagebackref=true,breaklinks=true,colorlinks,linkcolor=black,anchorcolor=black,citecolor=black,bookmarks=false]{hyperref}
\usepackage[capitalize]{cleveref}
\begin{document}

\title{Condition-Adaptive Graph Convolution Learning for Skeleton-Based Gait Recognition}

\author{Xiaohu~Huang,
        Xinggang~Wang, ~\IEEEmembership{Member,~IEEE},
        Zhidianqiu~Jin,
        Bo~Yang,
        Botao~He,
        ~Bin~Feng,
        and Wenyu~Liu, ~\IEEEmembership{Senior Member,~IEEE,}
\thanks{Xiaohu~Huang, Xinggang~Wang, Zhidianqiu~Jin, Bin~Feng and Wenyu~Liu are with the School of Electronic Information and Communications, Huazhong University of Science and Technology, Wuhan 430074, China. Bo~Yang and Botao~He are with Wuhan FiberHome Digital Technology Co., Ltd.}
\thanks{X. Huang(e-mail:huangxiaohu@hust.edu.cn).}
\thanks{X. Wang(e-mail:xgwang@hust.edu.cn).}
\thanks{Z. Jin(e-mail:jzdq@hust.edu.cn).}
\thanks{B. Yang(email: byang@fhzz.com.cn)}
\thanks{B. He(e-mail:hebotao@fhzz.com.cn)}
\thanks{B. Feng (e-mail: fengbin@hust.edu.cn). Corresponding author.}
\thanks{W. Liu(e-mail: liuwy@hust.edu.cn).}
}

\markboth{Journal of \LaTeX\ Class Files,~Vol.~14, No.~8, August~2021}%
{Shell \MakeLowercase{\textit{et al.}}: A Sample Article Using IEEEtran.cls for IEEE Journals}

\IEEEoverridecommandlockouts
\IEEEpubid{\makebox[\columnwidth]{978-1-5386-5541-2/18/\$31.00~\copyright2023 IEEE \hfill} \hspace{\columnsep}\makebox[\columnwidth]{ }}

\maketitle
\IEEEpubidadjcol
\begin{abstract}
Graph convolutional networks have been widely applied in skeleton-based gait recognition. A key challenge in this task is to distinguish the individual walking styles of different subjects across various views. Existing state-of-the-art methods employ uniform convolutions to extract features from diverse sequences and ignore the effects of viewpoint changes. To overcome these limitations, we propose a condition-adaptive graph (\textbf{CAG}) convolution network that can dynamically adapt to the specific attributes of each skeleton sequence and the corresponding view angle. In contrast to using fixed weights for all joints and sequences, we introduce a joint-specific filter learning (\textbf{JSFL}) module in the CAG method, which produces sequence-adaptive filters at the joint level. The adaptive filters capture fine-grained patterns that are unique to each joint, enabling the extraction of diverse spatial-temporal information about body parts. Additionally, we design a view-adaptive topology learning (\textbf{VATL}) module that generates adaptive graph topologies. These graph topologies are used to correlate the joints adaptively according to the specific view conditions. Thus, CAG can simultaneously adjust to various walking styles and viewpoints. Experiments on the two most widely used datasets (i.e., CASIA-B and OU-MVLP) show that CAG surpasses all previous skeleton-based methods. Moreover, the recognition performance can be enhanced by simply combining CAG with appearance-based methods, demonstrating the ability of CAG to provide useful complementary information. The source code will be available at \url{https://github.com/OliverHxh/CAG}
\end{abstract}


\section{Introduction}
\IEEEPARstart{G}{ait} recognition is an important biometric technology with various applications ranging from case detection to human-robot interaction. The main idea is to identify a person by his/her distinctive walking style. The gait recognition methods can be classified into two categories; appearance-based \cite{gaitset,gaitpart,CSTL,local3DCNN,GaitGL} and model-based ~\cite{PoseGait,gaitgraph,jointsgait,hmrgaitaccv,hmrgaiticcv,OU-MVLP}. Since the appearance-based methods are more sensitive to the appearance variations, model-based methods have gained more attention recently. Among the various model-based methods, the skeleton representation is the most popular because it can be easily used to extract features, and it is consistent with the human body structure.

\begin{figure}[t]
    \centering
    \includegraphics[width=0.9\linewidth]{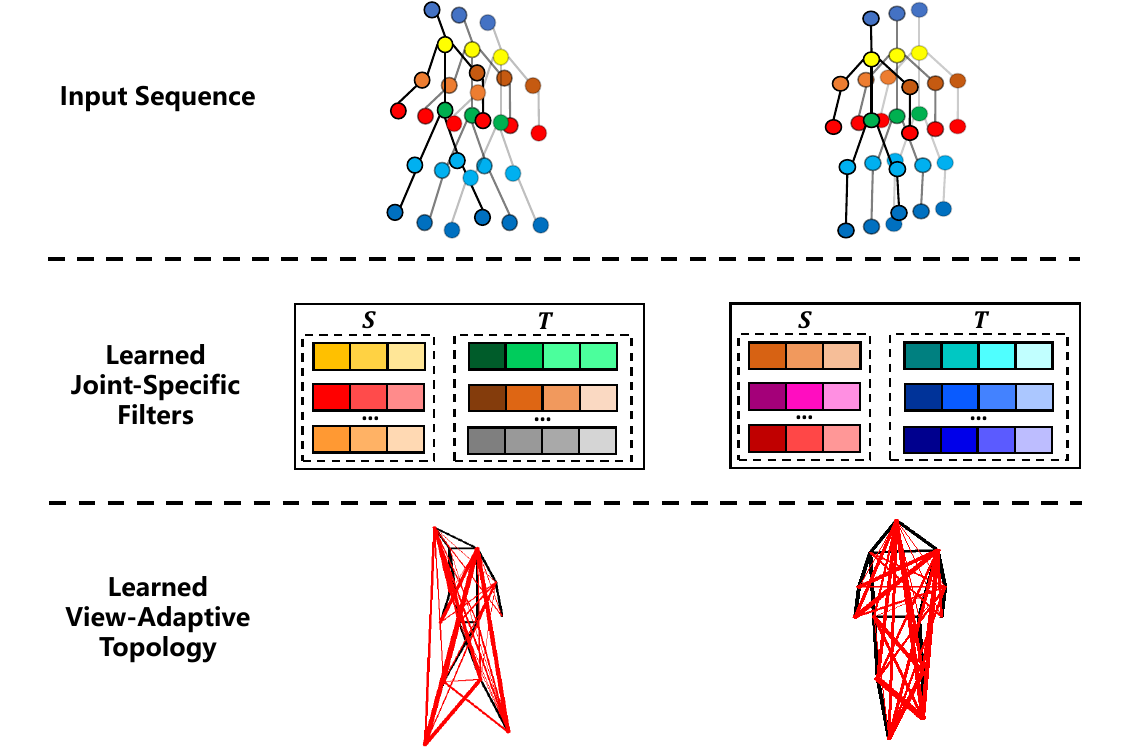}
    \caption{An overview of the proposed idea. \textbf{First row}: CAG takes gait-skeleton sequences as inputs. \textbf{Second row}: CAG automatically generates joint-specific filters for each sequence in both the \textbf{$S$} (spatial) and \textbf{$T$} (temporal) domains; thus, it can capture personalized walking styles and extract fine-grained patterns. \textbf{Third row}: CAG dynamically learns a view-adaptive topology for each sequence to handle customized gait characteristics across different camera views (Best viewed in color).}
    \label{fig:illustration}
\end{figure}

Graph convolutional networks (GCNs) have been widely applied to achieve impressive results in skeleton-based gait recognition \cite{gaitgraph, jointsgait,3DGCN, msgg, cyclegait, gaittake} since they can model inherent correlations between joints. These methods have used standard weight-sharing convolutions to extract the spatial and temporal features of each joint from the sequences. However, personalized gait characteristics exhibit complex patterns of joints. Thus, such uniform feature extraction can only capture a general human walking style but cannot adapt to individual walking styles. Considering that gait is a fine-grained motion pattern, it is essential to distinguish a personalized walking style among different individuals. Therefore, the fixed filters used in the above-mentioned methods limit the flexible and robust modeling ability.

Moreover, the used 2D skeleton structure shows different characteristics for different camera views, making the corresponding topological correlations diverse. Also, current gait methods \cite{gaitgraph,jointsgait,3DGCN} employ predefined graph topologies, which may not be suitable for all views. Some action recognition methods \cite{contextGCN,dynamicGCN,decoupledGCN} proposed adaptive graphs by learning the correlations between joints dynamically. However, such adaptive graphs were not designed to fit cross-view scenarios. Therefore, current skeleton-based methods do not offer explicit solutions to viewpoint variations. As a consequence, the recognition performance is hindered.

To tackle the above issues, we propose a novel GCN for skeleton-based gait recognition, called condition-adaptive graph (\textbf{CAG}) convolution network. The main idea of CAG is to adapt graph convolution learning to suit the variations in personalized walking styles and viewing conditions. As shown in \cref{fig:illustration}, in CAG, filters are automatically learned to capture personalized walking characteristics using a joint-specific filter learning (\textbf{JSFL}) module, and graphs that can handle viewpoint variations are generated using a view-adaptive topology learning (\textbf{VATL}) module. These dynamic filters are used to extract fine-grained spatial-temporal patterns of each joint, and the graphs are employed to correlate the joints adaptively, based on the specific viewing condition. Therefore, the JSFL and VATL modules can be seamlessly integrated together in the proposed network.

Specifically, the JSFL module produces filters by encoding joint-level features across the entire sequence. In particular, since different joints correspond to different body parts with various patterns, the network can exploit joint-level feature mining to obtain fine-grained information. The JSFL module constructs two branches, which correspond to model spatial configuration and temporal motion, respectively. This architecture enables the spatial and temporal filters to learn separately. Considering the computational efficiency, all filters are learned in a depth-wise manner.

The VATL module generates view-adaptive topologies by using prior-view knowledge. Specifically, VATL transforms the general \textbf{fixed view-invariant} topology into a set of \textbf{learnable view-related} topologies. It then constructs the view-adaptive topology with the following three components: (1) A topology, which is the most appropriate for the sequence view. (2) A topology, which is a weighted summation of all learnable view-related topologies, enhances its robustness by utilizing the intrinsic correlation of different views. (3) A fixed topology, which represents prior knowledge about the human body structure and has been shown effective in human action recognition \cite{STGCN,twostream,dynamicGCN}. In this way, the learned view-adaptive topology considers specific viewing conditions and incorporates general knowledge about the human body structure.

In summary, the main contributions of this paper include the following three aspects: \begin{enumerate} 
\item[(1)] A JSFL module that dynamically generates joint-specific filters tailored to sequence characteristics. In this way, graph convolutions can adapt to personalized walking styles and detailed spatial-temporal patterns can be extracted from each sequence. 
\item[(2)] A VATL module that generates a view-adaptive topology, based on specific viewing conditions in each sequence. In this way, graph convolutions can handle the view variations. 
\item[(3)] A condition-adaptive graph (CAG) convolutional network is proposed by integrating the JSFL and VATL modules. Extensive experiments conducted on CASIA-B \cite{CAISA-B} and OU-MVLP \cite{OU-MVLP} datasets demonstrate the state-of-the-art performance of CAG. By combining CAG with appearance-based methods, the recognition performance can be effectively improved. \end{enumerate}

\section{Related Research Work}
\subsection{Gait Recognition}
Currently, two categories of mainstream gait recognition methods are available; the \textbf{appearance-based} and the \textbf{model-based} methods. The \textbf{appearance-based} approaches obtain silhouettes as inputs, which rely on abundant shape information to model spatial-temporal features.

Some of the representative appearance-based methods are disentanglement-based, set-based, part-based, and 3D convolutional neural networks (CNNs)-based. The disentanglement-based methods \cite{GaitDistangle1,GaitDistangle2, disentanglepami} aimed to disentangle the original walking features into identity-relevant features and identity-irrelevant features, which avoided the negative effects of confounding variables. The set-based approaches \cite{gaitset,GLN} regarded a gait sequence as an unordered set, which processed each frame independently, and did not explicitly model temporal relations. Further, the part-based methods \cite{gaitpart,Condition-aware,CSTL, STAR} proposed to extract features of different parts individually for fine-grained feature extraction, and applied temporal motion modeling in different scales. 3D CNN-based methods \cite{3DCNN,MT3D,local3DCNN,GaitGL} stacked layers of 3D convolutions to capture spatial-temporal patterns in multiple scales.

The \textbf{model-based} approaches methods model the human structure and body movement by designing simulated models \cite{nixon1999automatic,wang2004fusion} or using skeletons \cite{PoseGait,OU-MVLP,jointsgait,gaitgraph} as inputs. Recently, due to the successful development of pose estimation methods \cite{hrnet,openpose,Alphapose}, the skeleton-based methods have prevailed.

The PoseGait \cite{PoseGait}, CNN-pose \cite{OU-MVLP}, and pose-based temporal–spatial network (PTSN) \cite{PTSN} methods used a skeleton sequence as a 2D matrix and employ 2D CNNs or LSTMs to model gait features. These methods did not consider the topological connections of the skeletons. The JointsGait \cite{jointsgait}, Mao et.al.~\cite{3DGCN}, GaitGraph \cite{gaitgraph}, MSGG \cite{msgg}, CycleGait \cite{cyclegait}, Gait-D \cite{gaitd} adopted GCN-based architectures from skeleton-based action recognition \cite{STGCN,ResGCN}. Recently, a transformer-based method \cite{gaittr} adopted transformer blocks to model the spatial and temporal correlations in a self-attention manner. Furthermore, the ModelGait \cite{hmrgaitaccv} and Li et.al.~\cite{hmrgaiticcv} methods used a human mesh-recovery (HMR) \cite{HMR} network to extract and use both shape and pose features.

The proposed CAG belongs to the \textbf{skeleton-based} methods and utilizes a GCN-based network architecture. 

\subsection{GCNs for Skeleton Modeling}
In recent years, numerous GCNs have been adopted to model spatial-temporal features in skeleton-based video analysis domains, especially in skeleton-based action recognition. Most current GCNs follow the pipeline design of ST-GCN \cite{STGCN}. For skeleton-based methods using GCNs (MSGG\cite{msgg}, CycleGait\cite{cyclegait}, GAITTAKE\cite{gaittake}, Gait-D\cite{gaitd}, GaitGraph\cite{gaitgraph}, and JointsGait\cite{jointsgait}), they process different sequences with the same network parameters in GCNs, therefore limiting the model capacity to extract sample-specific characteristics. On the contrary, the proposed JSFL module learns various filters for different sequences and joints, which benefit extracting personalized walking features. Besides, MSGG\cite{msgg} and GAITTAKE\cite{gaittake} adopt temporal attention approaches, which improve temporal aggregation flexibility. However, this adaptive manner is limited in the temporal domain and only used for feature aggregation, which does not play the main role in feature extraction, while our JSFL module is applied in both spatial and temporal domains, and used for feature extraction in the GCN backbone. A Transformer-based method (Gait-TR\cite{gaittr}) uses Transformer blocks to dynamically learn spatial gait patterns, but its temporal learning parameters are still shared for different samples, which is not flexible.

Some dynamic GCNs \cite{contextGCN,dynamicGCN,decoupledGCN} were proposed to learn joint correlations dynamically in order to relax the fixed topology constraints and enrich the global context. However, these methods were not designed to extract fine-grained features, and their graphs were not generated to relate explicitly to viewing conditions, which is crucial for gait recognition. In contrast, the proposed VATL module employs learning of adaptive topologies, explicitly based on viewing conditions.

\subsection{Adaptive Mechanisms}
Data-dependent mechanisms have achieved great success in computer vision, which adjust feature extractions to capture instance-specific properties. SE-Net \cite{SENET} connected the relations among different channels to adaptively attend to the most important ones. Self-attention methods \cite{attention,nonlocal,VIT} utilized QKV-based techniques to effectively construct the global context. Further, inspired by the attention ideas, the methods reported in \cite{condconv,dynamicconv,zhang2020dynet} generated dynamic weights to combine a set of filters in order to promote the network representation capacity. Recently, lightweight networks \cite{condinst,li2021involution,liu2021tam,ddf} produced convolutional filters on-the-fly, which adaptively fit the customized features.

For appearance-based methods (GaitPart\cite{gaitpart} and MetaGait\cite{metagait}) using attention mechanisms, their approaches are just supplements to the uniform feature extraction of their backbones. In contrast, the proposed JSFL module achieves dynamic feature extraction by generating adaptive convolutional filters, which no longer require an attention mechanism. Besides, the part-level feature learning in appearance-based methods is achieved by a manual partition, where the part semantics are not well aligned. In contrast, JSFL can obtain better-aligned parts from the skeleton inputs.

Previously, a few appearance-based gait approaches (MGAN\cite{mgan}, GaitGAN\cite{yu2017gaitgan}, Chai et.al.\cite{viewembedding}, and Vi-GaitGL\cite{viewembedding1}) have studied the topic of learning view-invariant gait features. MGAN\cite{mgan} and GaitGAN\cite{yu2017gaitgan}, and Makihara et.al \cite{viewtransform} transform gait energy images (GEIs), period energy images (PEIs) or silhouettes from arbitrary views into a targeted view, which however is not feasible for skeleton-based gait recognition. Chai et.al.\cite{viewembedding} and Vi-GaitGL\cite{viewembedding1} propose to learn view-specific embedding or projection parameters for the fully-connected layers. In contrast, the proposed view-adaptive topology learning (VATL) aims to generate view-adaptive topologies for GCNs.

\begin{figure*}[t]
    \centering
    \includegraphics[width=0.95\linewidth]{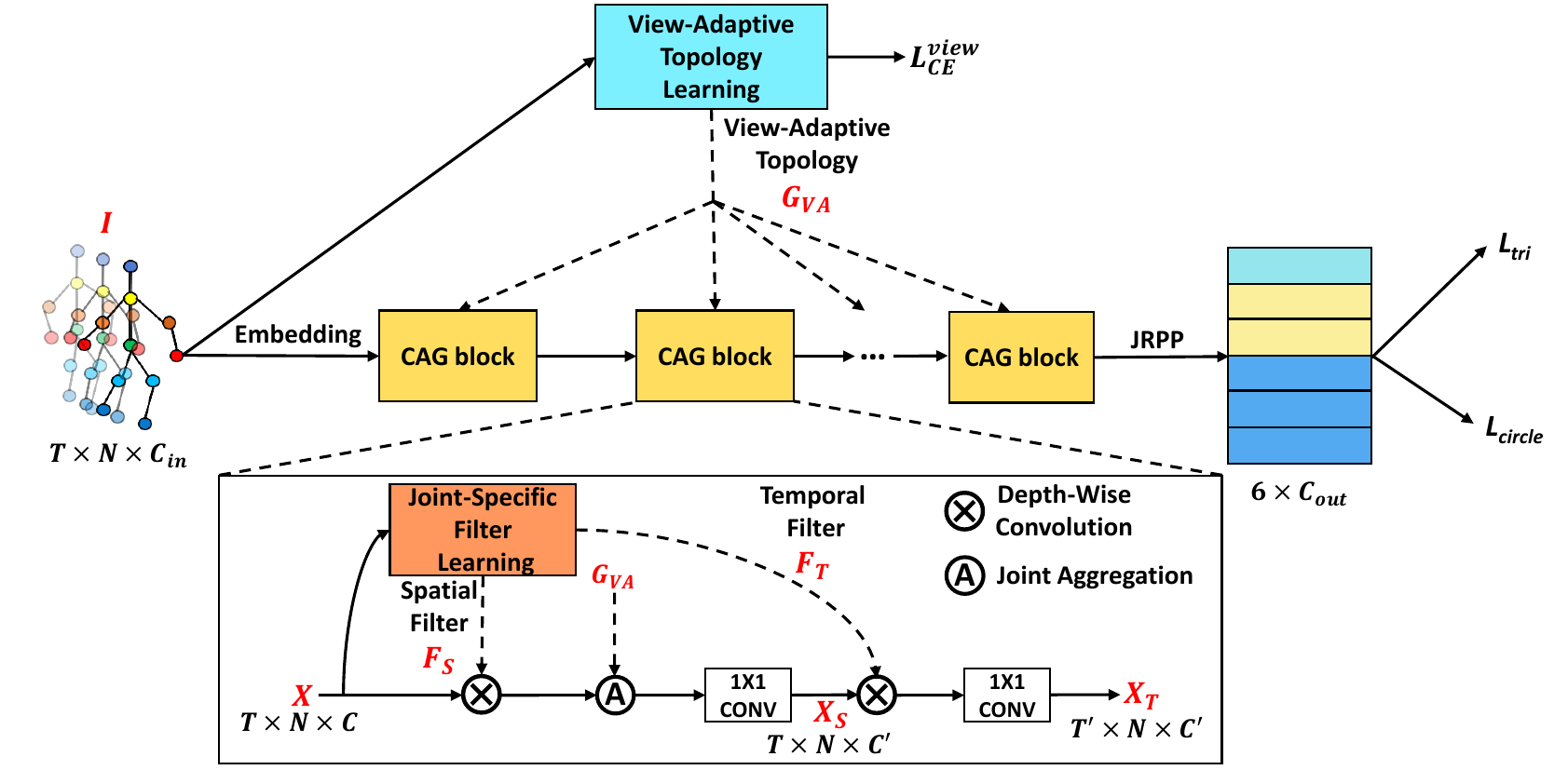}
    \caption{Illustration of CAG. A gait skeleton sequence of $T$ frames is taken as an input. The input sequence is passed through two branches; the first branch employs the VATL module to produce a view-adaptive topology for adaptation to camera viewpoints; the second branch extracts customized features using a series of CAG blocks. In each block, the adaptive joint-specific filters are dynamically generated by the JSFL module. Joints relationship pyramid mapping (JRPP) \cite{jointsgait} is used to map the gait skeleton features into 6 scales. $C_{in}$, $C_{out}$, $L_{CE}$, $L_{tri}$ and $L_{circle}$ denote the input-channel number, the output-channel number, the cross-entropy loss, the triplet loss \cite{tripletloss} and the circle loss \cite{circleloss}, respectively. See text for more details.}
    \label{fig:network}
\end{figure*}

\section{Proposed Method}
In this section, we initially review the preliminary concepts of GCN. Then, we describe the proposed network architecture and provide details of the proposed modules.

\subsection{Preliminary Concepts}
\noindent \textbf{Notations.} A human skeleton is denoted as a topology graph $\mathcal{G} = \{\mathcal{V}, \mathcal{E}\}$, where the vertex set $\mathcal{V}$ denotes body joints and the edge set $\mathcal{E}$ denotes bones. The vertex set is represented as $\mathcal{V}=\{v_1, v_2, ..., v_N\}$, where $N$ denotes the number of vertices. The edge set $\mathcal{E}$ is formulated as an adjacent matrix $A \in \mathbb{R}^{N \times N}$, where each element $a_{i,j}$ is defined as the connection strength between vertices $v_i$ and $v_j$. We formulate a skeleton sequence of $T$ frames as $I\in \mathbb{R}^{T \times N \times C_{in}}$, where $C_{in}$ denotes the input-channel dimension. For each input skeleton, we apply a batch normalization layer to normalize the joint coordinates before feeding them into the network. The features of the vertex set of $T$ frames are formulated as $X \in \mathbb{R}^{T \times N \times C}$, where $C$ denotes the channel dimension of each vertex. \\
\noindent \textbf{Graph Convolution.} Let $X_{S} \in \mathbb{R}^{T \times N \times C'}$ be the output features after performing spatial configuration extraction, where $C'$ denotes the output-channel dimension. In this way, the general graph convolution described in \cite{STGCN} follows the formulation given below:
\begin{equation}
    X_S = \sum_{k=1}^{K_S}{A^{k}XW^{k}_S},
    \label{eq:graph conv}
\end{equation}
where $K_S$ denotes the kernel size of the spatial domain (e.g., 3 in ST-GCN \cite{STGCN}), and $W_S \in \mathbb{R}^{K_S \times C \times C'}$ denotes the feature transformation filter in the spatial domain. The adjacent matrix $A^k \in \mathbb{R}^{N \times N}$ enables GCN to aggregate the information about vertices in a spatial context, which captures the human architecture configuration.

After the application of spatial configuration extraction, a kernel size of $K_T$ temporal convolution is employed by $W_T$ to model the temporal dynamics; the output $X_T$ with a temporal dimension $T'$ is obtained as follows:
\begin{equation}
    X_T = Conv(X_S, W_T),
    \label{eq:temporal conv}
\end{equation}
where $W_T \in \mathbb{R}^{K_T \times C' \times C'}$, and $X_T \in \mathbb{R}^{T' \times N \times C'}$.

\subsection{Network Architecture}
The two-branch architecture of CAG is illustrated in \cref{fig:network}, where a human skeleton sequence $I$ is taken as an input. In the first branch, based on the viewing conditions and sequence characteristics, VATL dynamically constructs a view-adaptive topology $G_{VA}$. The constructed $G_{VA}$ is utilized to guide a joint aggregation of graph convolutions in the second branch. Also, a cross-entropy loss $L_{CE}^{view}$ is employed to supervise view-related feature learning.

In the second branch, coarse skeleton features are initially extracted using a lightweight embedding module, i.e., an ordinary GCN block, which is described in \cref{eq:graph conv}. Then, CAG blocks are stacked to refine the features and extract customized clues. In each block, the JSFL module is used to automatically generate filters ($F_S$ and $F_T$) in a depth-wise manner. In particular, $F_S$ combined with $G_{VA}$ from VATL is used for spatial configuration extraction, and $F_T$ is used for temporal modeling. Two $1 \times 1$ convolutions are employed to fuse the cross-channel information. 

Consequently, the GCN operations in \cref{eq:graph conv} and \cref{eq:temporal conv} can be rewritten as follows:
\begin{equation}
    \begin{aligned}
        &X_S = Conv_{1 \times 1}(\sum_{k=1}^{K_S}{G_{VA}^{k}(X \otimes F^{k}_S}), W_1), \\
        &X_T = Conv_{1 \times 1}(X_S \otimes F_T, W_2),
    \end{aligned}
\end{equation}
where $\otimes$ denotes the depth-wise convolution, $G_{VA}^{k} \in \mathbb{R}^{N \times N}$, $F^{k}_S \in \mathbb{R}^{N \times C}$, $F_T \in \mathbb{R}^{K_T \times N \times C'}$, $W_1 \in \mathbb{R}^{C \times C'}$ and $W_2 \in \mathbb{R}^{C' \times C'}$.

After processing the CAG blocks, a JRPP \cite{jointsgait} module is used to map the gait features into 6 scales based on the joint relationship of human architecture. Finally, a triplet loss \cite{tripletloss} $L_{tri}$ and a circle loss \cite{circleloss} $L_{circle}$ are applied on the output features to perform training supervision. The overall loss function is summarized as follows:
\begin{equation}
    L = \lambda_1 L_{tri} + \lambda_2 L_{circle} + \lambda_3 L_{CE}^{view},
    \label{eq:loss}
\end{equation}
where $\lambda_1$, $\lambda_2$ and $\lambda_3$ are the hyperparameters to balance the respective loss functions.

Inspired by the two-stream works \cite{twostream,disentangleGCN}, where fusing joint and bone features enable networks to recognize human activities, we employ two separate streams for individually extracting joint and bone features; the objective is to combine the merits of both. The pipeline of the bone-feature stream is the same as that of the joint-feature stream, except that its input is the subtraction of coordinates in adjacent joints. The joint-feature stream can be regarded as the first-order information, and the bone-feature stream can be regarded as the second-order information. For simplicity, we only illustrate the joint-feature stream in \cref{fig:network}. Features from the two streams are concatenated with a size of $12 \times C_{out}$ and used as an output.

\begin{figure}
    \centering
    \includegraphics[width=0.8\linewidth]{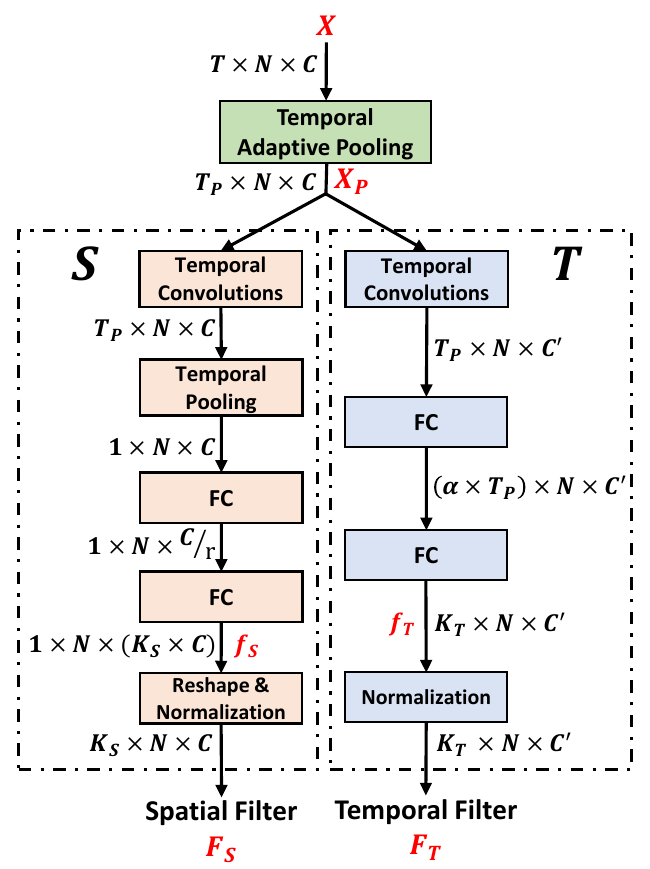}
    \caption{JSFL constructs two separate branches to generate $S$ (spatial) and $T$ (temporal) joint-specific filters, respectively.}
    \label{fig:JSFL}
\end{figure}

\subsection{Joint-Specific Filter Learning}
Since different body parts typically exhibit different amounts of variation and degrees of freedom due to the articulated structure of the skeleton, the JSFL module is used to describe the individual spatial-temporal characteristics flexibly in different gait sequences by generating customized filters. As shown in \cref{fig:JSFL}, two separate branches corresponding to spatial and temporal filter generations are utilized to extract the spatial configurations and capture the temporal dynamics, respectively. Particularly, the filters are generated in a depth-wise manner to increase efficiency. 

The gait features $X \in \mathbb{R}^{T \times N \times C}$ are used as an input and a temporal adaptive pooling is applied to obtain a temporal downsampled output $X_P \in \mathbb{R}^{T_P \times N \times C}$, where $T_P$ denotes the pooled size. For both spatial and temporal branches, we initially utilize temporal convolutions to learn the contextual information at each joint. Then, we apply a temporal pooling (TP) operation to the spatial branch to aggregate the temporal global context. Next, two cascaded fully-connected layers with a batch normalization layer and a rectified linear unit (ReLU) activation function are used to construct the cross-channel communications and produce filter $f_s$ with the expected size $1 \times N \times (K_S \times C)$. Subsequently, we reshape the filter into a size $K_S \times N \times C$, and adopt batch normalization to avoid filter parameters being extremely large or small. In summary, the operations in the spatial branch can be formulated as follows:
\begin{equation}
    \begin{aligned}
        &f_S = \mathcal{F}(\mathcal{F}(TP(TC(X_P)), W_3), W_4),\\
        &F_S = BN(Reshape(f_s)),
    \end{aligned}
\end{equation}
where $TC$, $TP$, $\mathcal{F}$ and $BN$ denote temporal convolution, temporal pooling, fully-connected layer, and batch normalization respectively. $W_3 \in \mathbb{R}^{C \times \frac{C}{r}}$ reduces the channel dimension by the ratio $r$ and $W_4 \in \mathbb{R}^{\frac{C}{r} \times C}$ recovers the channel dimension.

Different from the spatial branch, the temporal branch is used to describe motion characteristics, which does not include TP for maintaining the temporal structure, and uses two cascaded fully-connected layers with a ReLU activation function along the temporal dimension. The objective is to effectively exploit rich temporal relations in different moments in order to explore motion properties. Next, a normalization operation is applied to ensure parameter distribution stability. In summary, the operations in the temporal branch can be formulated as follows:
\begin{equation}
    \begin{aligned}
        &f_T = \mathcal{F}(\mathcal{F}(TC(X_P), W_5), W_6),\\
        &F_T = BN(f_T),
    \end{aligned}
\end{equation}
where $W_5 \in \mathbb{R}^{T_P \times (\alpha \times T_P)}$ inflates the temporal dimension by the ratio $\alpha$ and $W_6 \in \mathbb{R}^{(\alpha \times T_P) \times K_T}$ reduces it to a defined size $K_T$.

\noindent \textbf{Discussion.} Some appearance-based gait methods \cite{gaitpart, GaitGL, CSTL} use part-based approaches to model the local features. These methods are similar to JSFL to some extent. We compare JSFL with these part-based approaches. Their differences are summarized as follows: a) The part-based methods extract features of different sequences using uniform convolutions, whereas JSFL extracts features for each sequence adaptively. b) The part-based methods use non-shared convolutions, whose parameter usage is times larger than that of the vanilla convolutions. However, JSFL, which saves approximately half of the parameters shown in the first and second rows of \cref{tab:module effectiveness}, is more efficient than the vanilla convolution. c) The part-based methods mostly obtain the parts using a manual partition, where the part semantics are not well aligned. In contrast, JSFL can obtain the well-aligned parts from the skeleton inputs.

\begin{figure}[t]
    \centering
    \includegraphics[width=0.9\linewidth]{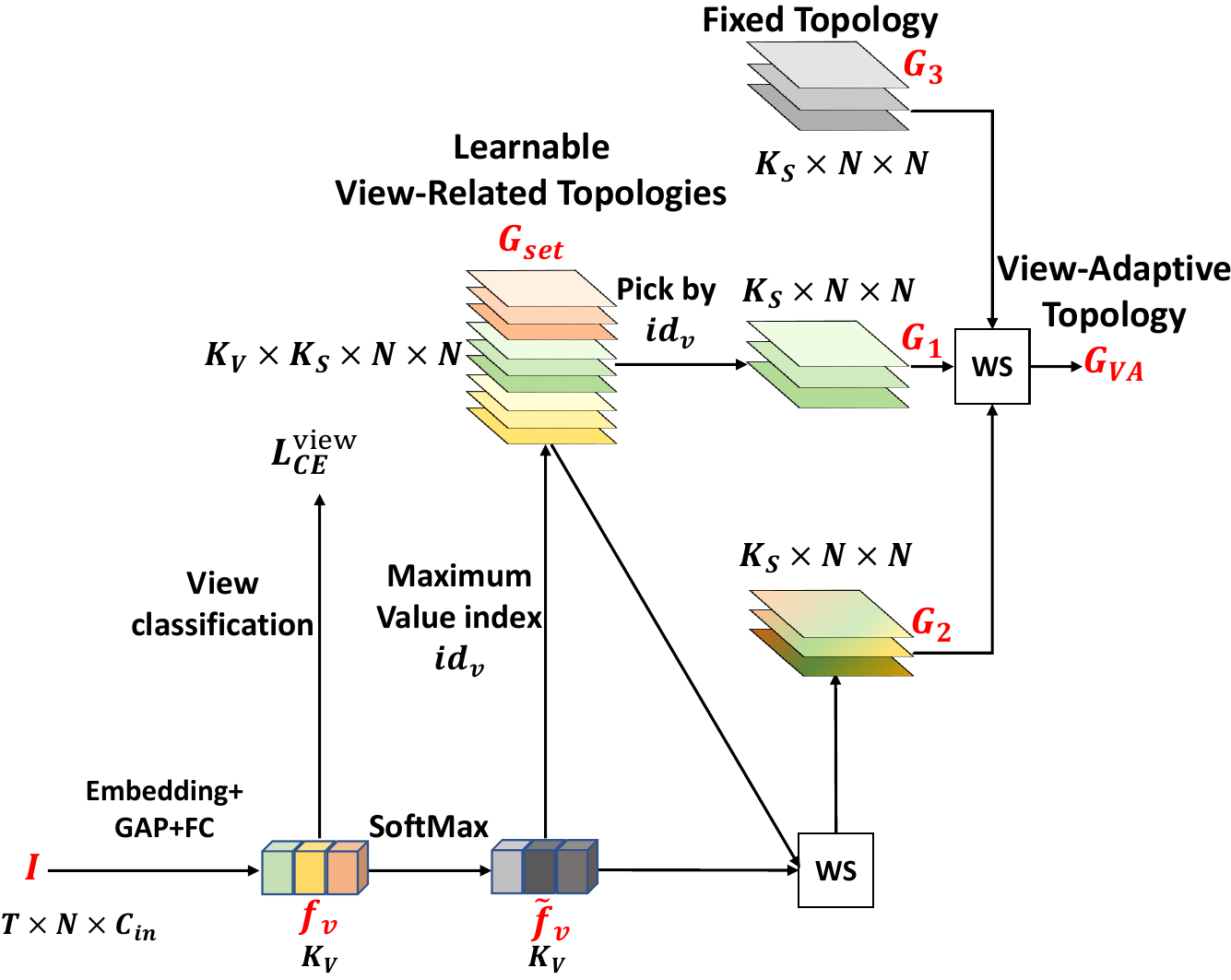}
    \caption{VATL defines a set of learnable view-related topologies and employs prior-view knowledge to learn the view-adaptive topology. GAP, FC, and WS denote global average pooling, fully-connected layer, and weighted summation respectively.}
    \label{fig:VATL}
\end{figure}

\subsection{View-Adaptive Topology Learning}
The VATL module utilizes the intrinsic view information in each sequence to learn a view-adaptive topology. As shown in \cref{fig:VATL}, the original gait sequence $I$ is obtained as an input. First, we apply an embedding module to extract view-related features, and then use GAP to aggregate the global view information. Next, a fully-connected layer is employed to obtain the view-classification vector $f_v \in \mathbb{R}^{K_V}$, where $K_V$ denotes the number of views (e.g., 11 in CAISA-B \cite{CAISA-B} and 14 in OU-MVLP \cite{OU-MVLP}). Here, the cross-entropy loss on $f_v$, which produces a loss $L_{CE}^{view}$, is used to supervise feature learning. This ensures view prediction ability. Subsequently, a SoftMax function is employed to produce a value-normalized vector $\widetilde{f}_{v} \in \mathbb{R}^{K_V}$. 

A set of learnable topologies $G_{set}=\{G_{V}^1, G_{V}^2,...,G_{V}^{K_V}\}$ is defined to enhance the view-adaptive capacity, where $G_{V}^i \in \mathcal{R}^{K_S  \times N \times N}$ denotes the corresponding topology obtained from the $i$-th view. To generate the view-adaptive topology $G_{VA}$, we first obtain the index of the maximum value in $\widetilde{f}_{v}$, which is formulated as:
\begin{equation}
    id_v = \arg\max{\widetilde{f}_{v}},
\end{equation}
where $id_v$ indicates the view that the sequence most possibly encounters. Then, the corresponding topology is selected from $G_{set}$ by $id_v$ as follows:
\begin{equation}
    G_1 = G_{set}[id_v],
\end{equation}
where $G_1$ reflects the particular properties that the corresponding view possesses. On the test sets of CASIA-B and OU-MVLP datasets, we achieve top-1 view-classification accuracies of 98.5\% and 98.7\%, respectively, which are quite reliable. Therefore, for each sequence, given the predicted view-classification result, we can accurately select the adaptive topology for the corresponding view. Considering that the topology is initialized as learnable parameters, it can be updated to adapt to the view characteristics through the backward propagation technique. However, $G_1$ is not sufficient to represent all types of intra-variation existing in this view; thus, we introduce a supplementation. Considering that the data distribution in $\widetilde{f}_v$ reveals the sequence characteristics to some extent, we consider the data as linear weights to combine the topologies in $G_{set}$. This can be formulated as follows:
\begin{equation}
    G_2 = \sum_{i=1}^{K_V}{\widetilde{f}_{v}^{i}G_{V}^i},
\end{equation}
Also, we use a fixed topology $G_3$, which is an ordinary graph in \cref{eq:graph conv} to extract the general feature representation. Finally, we fuse $G_1$, $G_2$ and $G_3$ with coefficients $g_1$, $g_2$ and $g_3$ to obtain $G_{VA}$:
\begin{equation}
    G_{VA} = g_1 G_1 + g_2 G_2 + g_3 G_3.
\end{equation}
$G_{VA}$ is used to dynamically connect body joints in the graph convolutions, which not only correlate joints in the nearby locations but also incorporate joint information in the long range. The dynamic complex connections can effectively enhance the adaptation ability in cross-view scenarios.

\section{Experiments}
\subsection{Datasets}
\noindent \textbf{CASIA-B.} The CASIA-B \cite{CAISA-B} dataset contains 124 walking subjects, and each subject includes 110 sequences obtained from 11 camera views. For each view, each subject has 10 sequences of 3 walking conditions, i.e., 6 sequences of normal (NM) walking, 2 sequences of walking with a bag (BG), and 2 sequences of walking with a coat (CL). The training and testing settings followed the protocols reported in \cite{gaitcomprehensive}. The sequences of the first 74 subjects were used for training, and the sequences of the remaining 50 subjects were used for testing. Specifically, in the testing phase, the NM sequences were used as gallery sets, and the BG and CL sequences were used as probe sets. The skeleton data were extracted by HRNet \cite{hrnet} and OpenPose \cite{openpose}. Unless otherwise stated, the HRNet data were used on CASIA-B.

\noindent \textbf{OU-MVLP.} The OU-MVLP \cite{OU-MVLP} dataset contains 10307 subjects, and each subject includes 28 sequences obtained from 14 camera views. For each view, each subject has 2 sequences (index '01' and index '02'). The training and testing followed the protocols reported in \cite{OU-MVLP}. The sequences of the first 5153 subjects were used for training, and the sequences of the remaining 5154 subjects were used for testing. Specifically, in the testing phase, the sequences with index '01' were used as gallery sets, and the sequences with index '02' were used as probe sets. This dataset provides skeleton data estimated by AlphaPose \cite{Alphapose} and OpenPose \cite{openpose}. In this paper, we used skeleton data extracted by AlphaPose \cite{Alphapose}.

\begin{table*}[t]
\begin{center}
\caption{Comparison of the proposed CAG method with skeleton-based methods using the CASIA-B dataset in term of the averaged rank-1 accuracies (\%), excluding identical-view cases. * stands for using 5 modalities as inputs as the same as Gait-TR \cite{gaittr}.}
\begin{tabular}{c|c|c| c c c c c c c c c c c|c}
\toprule[1.2pt]
\multicolumn{2}{c|}{Gallery NM} & Pose & \multicolumn{11}{|c|}{$0-180^\circ$} & \multirow{2}{*}{Avg} \\\cline{1-14}
\multicolumn{2}{c|}{Probe} & $-$ & $0^\circ$ & $18^\circ$ & $36^\circ$ & $54^\circ$ & $72^\circ$ & $90^\circ$ & $108^\circ$ & $126^\circ$ & $144^\circ$ & $162^\circ$ & $180^\circ$ & \\\hline
\multirow{9}{*}{NM} & {PoseGait \cite{PoseGait}} & 3D Pose & 55.3 & 69.6 & 73.9 & 75.0 & 68.0 & 68.2 & 71.1 & 72.9 & 76.1 & 70.4 & 55.4 & 68.7 \\\cline{2-15}
& JointsGait \cite{jointsgait} & OpenPose & 68.1 & 73.6 & 77.9 & 76.4 & 77.5 & 79.1 & 78.4 & 76.0 & 69.5 & 71.9 & 70.1 & 74.4 \\\cline{2-15}
& GaitGraph \cite{gaitgraph} & HRNet & 85.3 & 88.5 & 91.0 & 92.5 & 87.2 & 86.5 & 88.4 & 89.2 & 87.9 & 85.9 & 81.9 & 87.7 \\\cline{2-15}
& ModelGait \cite{hmrgaitaccv} (pose\_CNN) & HMR & 87.1 & 88.3 & 93.8 & 95.4 & 92.1 & 92.8 & 90.5 & 90.7 & 88.5 & 92.4 & 91.7 & 91.2 \\\cline{2-15}
& Li. et.al. \cite{hmrgaiticcv} (pose) & HMR &-&-& -&- &- &- &- & -& -& -& -& 93.1 \\\cline{2-15}
& MSGG \cite{msgg} & HRNet & 88.8 & 92.6 & 84.2 & 94.0 & 93.0 & 93.9 & 92.3 & 94.5 & 94.4 & 94.9 & 90.9 & 93.0 \\\cline{2-15}
 & CycleGait \cite{cyclegait} & HRNet & 92.3 & 93.2 & 92.9 & 93.9 & 91.9 & 94.1 & 94.3 & 93.3 & 92.8 & 91.1 & 91.1 & 92.8 \\\cline{2-15}
 & Gait-D \cite{gaitd} & HRNet & 87.7 & 92.5 & 93.6 & 95.7 & 93.3 & 92.4 & 92.8 & 93.4 & 90.6 & 88.6 & 87.3 & 91.6 \\\cline{2-15}
& Gait-TR \cite{gaittr} & HRNet & 95.7 & 96.4 & \textbf{97.9} & 97.0 & 96.9 & 95.5 & \textbf{95.1} & 96.1 & \textbf{96.6} & 96.0 & 92.4 & 96.0 \\\cline{2-15}
& \multirow{2}{*}{\textbf{CAG} (proposed)} & OpenPose & 90.5 & 91.8 & 94.1 & 94.3 & 94.3 & 92.3 & 93.2 & 92.1 & 93.1 & 91.2 & 87.7 & 92.2 \\\cline{3-15}
&  & HRNet & 94.2 & 96.3 & 96.8 & 96.2 & 96.2 & 96.0 & 94.8 & 96.8 & 96.4 & 96.5 & 93.0 & 95.7 \\\cline{2-15} 
& \textbf{CAG} (proposed)* & HRNet & \textbf{96.3} & \textbf{96.5} &  97.8 & \textbf{97.3} & \textbf{97.2} & \textbf{96.4} & \textbf{95.1} & \textbf{97.2} & \textbf{96.6} & \textbf{96.7} & \textbf{93.5} & \textbf{96.4} \\\midrule[1.2pt]

\multirow{9}{*}{BG} & {PoseGait \cite{PoseGait}} & 3D Pose & 35.3 & 47.2 & 52.4 & 46.9 & 45.5 & 43.9 & 46.1 & 48.1 & 49.4 & 43.6 & 31.1 & 44.5 \\\cline{2-15}
& JointsGait \cite{jointsgait} & OpenPose & 54.3 & 59.1 & 60.6 & 59.7 & 63.0 & 65.7 & 62.4 & 59.0 & 58.1 & 58.6 & 50.1 & 59.1 \\\cline{2-15}
& GaitGraph \cite{gaitgraph} & HRNet & 75.8 & 76.7 & 75.9 & 76.1 & 71.4 & 73.9 & 78.0 & 74.7 & 75.4 & 75.4 & 69.2 & 74.8 \\\cline{2-15}
& ModelGait \cite{hmrgaitaccv} (pose\_CNN) & HMR & 86.8 & 81.2 & 84.6 & 86.8 & 84.9 & 83.0 & 83.9 & 82.8 & 82.1 & 84.0 & 83.2 & 83.9 \\\cline{2-15}
& Li. et.al. \cite{hmrgaiticcv} (pose) & HMR &-&-& -&- &- &- &- & -& -& -& -& 88.0 \\\cline{2-15}
& MSGG \cite{msgg} & HRNet & 77.9 & 81.3 & 81.7 & 80.2 & 78.2 & 73.8 & 76.5 & 77.0 & 78.6 & 80.5 & 73.0 & 78.1 \\\cline{2-15}
& CycleGait \cite{cyclegait} & HRNet & 87.3 & 85.5 & 85.0 & 84.1 & 82.3 & 82.9 & 84.6 & 82.7 & 81.7 & 85.6 & 82.4 & 84.0 \\\cline{2-15}
& Gait-D \cite{gaitd} & HRNet & 78.2 & 80.1 & 79.3 & 80.2 & 78.4 & 77.6 & 80.4 & 78.6 & 79.1 & 80.2 & 76.5 & 79.0 \\\cline{2-15}
& Gait-TR \cite{gaittr} & HRNet & \textbf{90.9} & \textbf{92.4} & 91.4 & \textbf{93.2} & \textbf{91.9} & 90.2 & 91.4 & \textbf{93.9} & \textbf{93.9} & 92.7 & 82.9 & 91.3 \\\cline{2-15}
& \multirow{2}{*}{\textbf{CAG} (proposed)} & OpenPose & 81.5 & 86.9 & 88.8 & 86.8 & 85.6 & 84.1 & 85.9 & 86.6 & 85.1 & 83.3 & 75.0 & 84.5 \\\cline{3-15}
&  & HRNet & 87.5 & 90.0 & 91.0 & 91.1 & 87.8 & 88.1 & 89.2 & 91.8 & 90.6 & 91.6 & 87.1 & 89.6 \\\cline{2-15} 
& \textbf{CAG} (proposed)* & HRNet & 90.5 & 92.0 & \textbf{91.9} & 92.8 & 91.0 & \textbf{90.4} & \textbf{91.7} & \textbf{93.9} & 93.6 & \textbf{92.8} & \textbf{89.1} & \textbf{91.8} \\\midrule[1.2pt]

\multirow{9}{*}{CL} & {PoseGait \cite{PoseGait}} & 3D Pose & 24.3 & 29.7 & 41.3 & 38.8 & 38.2 & 38.5 & 41.6 & 44.9 & 42.2 & 33.4 & 22.5 & 36.0 \\\cline{2-15}
& JointsGait \cite{jointsgait} & OpenPose & 48.1 & 46.9 & 49.6 & 50.5 & 51.0 & 52.3 & 49.0 & 46.0 & 48.7 & 53.6 & 52.0 & 49.8 \\\cline{2-15}
& GaitGraph \cite{gaitgraph} & HRNet & 69.6 & 66.1 & 68.8 & 67.2 & 64.5 & 62.0 & 69.5 & 65.6 & 65.7 & 66.1 & 64.3 & 66.3 \\\cline{2-15}
& ModelGait \cite{hmrgaitaccv}(pose\_CNN) & HMR & 63.0 & 62.4 & 66.3 & 65.2 & 61.9 & 58.2 & 58.3 & 59.1 & 56.8 & 55.4 & 55.6 & 60.2 \\\cline{2-15}
& Li. et.al. \cite{hmrgaiticcv} (pose) & HMR &-&-& -&- &- &- &- & -& -& -& -& 64.3\\\cline{2-15}
& MSGG \cite{msgg} & HRNet & 62.2 & 67.4 & 66.2 & 70.2 & 68.8 & 66.2 & 67.4 & 96.2 & 71.1 & 73.4 & 69.7 & 68.3 \\\cline{2-15}
& CycleGait \cite{cyclegait} & HRNet & 78.6 & 76.8 & 79.2 & 80.5 & 78.0 & 77.6 & 81.2 & 77.1 & 76.5 & 82.4 & 77.7 & 78.7 \\\cline{2-15}
& Gait-D \cite{gaitd} & HRNet & 73.2 & 71.7 & 75.4 & 73.2 & 74.6 & 72.3 & 74.1 & 70.5 & 69.4 & 71.2 & 66.7 & 72.0 \\\cline{2-15}
& Gait-TR \cite{gaittr} & HRNet & \textbf{86.7} & 88.2 & 88.4 & 89.7 & \textbf{91.1} & 90.7 & \textbf{93.2} & \textbf{93.8} & \textbf{93.2} & \textbf{91.2} & 83.6 & 90.0 \\\cline{2-15}
& \multirow{2}{*}{\textbf{CAG} (proposed)} & OpenPose & 67.1 & 73.1 & 80.1 & 77.6 & 80.6 & 79.4 & 78.8 & 75.1 & 78.3 & 73.4 & 69.5 & 75.7 \\\cline{3-15}
&  & HRNet & 84.1 & 87.3 & 88.6 & 90.3 & 89.9 & 89.6 & 91.0 & 90.6 & 89.6 & 89.3 & 86.9 & 88.6 \\\cline{2-15} 
& \textbf{CAG} (proposed)* & HRNet & 86.3 & \textbf{88.8} & \textbf{89.3} & \textbf{91.0} & \textbf{91.1} & \textbf{91.2} & 93.0 & 92.8 & 92.6 & 91.0 & \textbf{89.0} & \textbf{90.6} \\\bottomrule[1.2pt]
\end{tabular}
\label{tab:casia-b}
\end{center}
\end{table*}

\subsection{Implementation Details}
\noindent \textbf{Hyperparameters.} The detailed hyperparameters applied to CASIA-B and OU-MVLP datasets are listed in \cref{tab:hyper-parameter}.
\begin{table}[h]
    \centering
    \caption{Hyperparameter settings applied to CASIA-B / OU-MVLP datasets.}
    \begin{tabular}{c|c|c|c|c|c|c}
     \toprule[1.2pt]
      $T$ & $N$ & $K_S$ & $K_T$ & $K_V$ & $T_P$ & $r$ \\\hline
      60/32 & 17/18 & 3/3 & 9/9 & 11/14 & 15/15 & 8/8 \\\hline
      $\alpha$ & $g_1$ & $g_2$ & $g_3$ & $\lambda_1$ & $\lambda_2$ & $\lambda_3$ \\\hline
      2/2 & $\frac{1}{2}$/$\frac{1}{2}$ & $\frac{1}{2}$/$\frac{1}{2}$ & 1/1 & 0.9/0.9 & 0.1/0.1 & 0.1/0.1 \\\bottomrule[1.2pt]
    \end{tabular}
    \label{tab:hyper-parameter}
\end{table}

\noindent \textbf{Training Details.} 1) The batch size in the training phase was set to $(p,k)$, where $p$ denotes the number of subjects and $k$ denotes the number of sequences for each subject. The batch sizes used in CASIA-B and OU-MVLP were $(8,16)$ and $(32, 12)$, respectively. 2) The margins in the triplet loss and circle loss were set to $0.2$ and $0.5$ respectively. 3) Since the data amount of OU-MVLP is twenty times larger than that of CASIA-B, the number of output channels in the embedding module was set as follows: 5-layer stacked CAG blocks and fully-connected layers to 64/128, 128/256, 128/256, 256/512, 256/512, and 256/512 in CASIA-B and OU-MVLP, respectively. 4) In total, 500 epochs were trained by Adam optimizer. The initial learning rate was set to 1e-4 for the VATL module and to 1e-3 for the remaining parameters. The learning rates were iteratively scaled by the step LR decay with a ratio of 0.1 at 255, 355, and 455 epochs. Also, a warmup strategy for the first 5 epochs was adopted to achieve increased stability in the training process.

\begin{table}[t]
    \centering
    \caption{Comparison of the proposed CAG method with skeleton-based methods using the OU-MVLP dataset; averaged rank-1 accuracies (\%), excluding identical-view cases; * indicates that the results are produced by the authors.}
    \setlength{\tabcolsep}{0.7mm}{
    \begin{tabular}{c|c|c|c}
        \toprule[1.2pt]
        \multirow{2}{*}{Probe} & \multicolumn{3}{c}{Gallery All 14 views} \\\cline{2-4}
        & CNN-Pose \cite{PoseGait} & GaitGraph \cite{gaitgraph}* & \textbf{CAG (ours)} \\\hline
        $0^\circ$ & 12.3 & 24.4& \textbf{45.4} \\\hline
        $15^\circ$ & 22.7 & 36.8& \textbf{61.2} \\\hline   
        $30^\circ$ & 29.3 & 40.3& \textbf{64.7} \\\hline
        $45^\circ$ & 31.5 & 42.5& \textbf{67.6} \\\hline
        $60^\circ$ & 30.5 & 41.9& \textbf{67.0} \\\hline
        $75^\circ$ & 24.7 & 38.9& \textbf{63.5} \\\hline
        $90^\circ$ & 18.1 & 33.5& \textbf{57.7} \\\hline
        $180^\circ$ & 8.7 & 21.3& \textbf{39.9} \\\hline
        $195^\circ$ & 12.3 & 24.6& \textbf{48.3} \\\hline
        $210^\circ$ & 15.5 & 21.7& \textbf{44.0} \\\hline
        $225^\circ$ & 23.5 & 34.0& \textbf{61.0} \\\hline
        $240^\circ$ & 23.3 & 33.5& \textbf{60.8} \\\hline
        $255^\circ$ & 18.3 & 30.4& \textbf{57.1} \\\hline
        $270^\circ$ & 15.2 & 27.1& \textbf{52.1} \\\hline
        Avg & 20.4 & 30.4 & \textbf{56.4} \\\bottomrule[1.2pt]
    \end{tabular}}
    \label{tab:ou-mvlp}
\end{table}

\subsection{Comparison with State-of-the-art Methods}
\noindent \textbf{CASIA-B.} A performance comparison between the proposed method and skeleton-based methods for 3 walking conditions and 11 views on CASIA-B \cite{CAISA-B} is presented in \cref{tab:casia-b}. The following three main observations can be made: (1) CAG achieves the best performance in all three walking conditions, which proves strong feature representation ability. (2) CAG is robust with various pose estimation methods and achieves the best performance for both HRNet and OpenPose. 

\begin{table}[h]
    \centering
    \caption{Comparison of the proposed CAG method with skeleton-based action recognition methods using the CASIA-B dataset.}
    \begin{tabular}{c|c|c|c|c}
        \toprule[1.2pt]
        \multirow{2}{*}{Model} & \multicolumn{4}{c}{Rank-1 Accuracy} \\\cline{2-5}
        & NM & BG & CL & Avg \\\hline
        2S-AGCN \cite{twostream} & 92.7 & 80.8 & 79.3 & 84.2 \\\hline
        MSG3D \cite{disentangleGCN} & 92.0 & 81.4 & 80.1 & 84.5 \\\hline
        CTR-GCN \cite{channel-wise} & 92.3 & 80.6 & 76.7 & 83.2 \\\hline
        \textbf{CAG} & \textbf{95.7} & \textbf{89.6} & \textbf{88.6} & \textbf{91.3} \\\bottomrule[1.2pt]
    \end{tabular}
    \label{tab:action_skeleton_based}
\end{table}

Since skeleton-based action recognition is a similar application to skeleton-based gait recognition, some state-of-the-art action recognition methods are compared with CAG using the CASIA-B dataset (\cref{tab:action_skeleton_based}).  For a fair comparison, all methods adopted the two-stream architecture. As shown in \cref{tab:action_skeleton_based}, CAG outperforms all these methods, which proves the superiority of gait-specific designs in CAG.

\noindent \textbf{OU-MVLP.}  A performance comparison of the proposed CAG method against skeleton-based methods using AlphaPose data in the OU-MVLP dataset is shown in \cref{tab:ou-mvlp}. CAG outperforms the current approaches marginally for all camera views, which further demonstrates its feature representation generality in a large-scale dataset and robustness in cross-view scenarios.

\subsection{Comparison with Appearance-based Methods}
In this section, we compare CAG with appearance-based methods using the CASIA-B dataset in terms of rank-1 accuracy, computational cost, and inference speed in \cref{tab:comparison_appearance_based_methods}. The computational cost and inference speed are measured using a 100-frame sequence for all models. The advantages of CAG include the following two main aspects: 1) CAG achieves higher performances in the CL condition than appearance-based methods, which indicates its stronger robustness against clothing variations. 2) CAG is more computationally efficient during training and inference phases.

\begin{table}[h]
    \centering
    \caption{Comparison of the proposed CAG method with appearance-based methods using the CASIA-B dataset.}
    \begin{tabular}{c|c|c|c|c|c}
    \toprule[1.2pt]
        \multirow{2}{*}{Model}& \multicolumn{3}{c|}{Rank-1 Accuracy(\%)} & \multirow{2}{*}{GFLOPs} & Inference \\\cline{2-4}
        & NM & BG & CL & & Time (ms) \\\hline
        GaitSet \cite{gaitset} & 95.0 & 87.2 & 70.4 & 21.4 & 2.4 \\\hline
        GaitPart \cite{gaitpart} & \textbf{96.2} & \textbf{91.5} & 78.7 & 21.4 & 4.9 \\\hline
        \textbf{CAG} & 95.7 & 89.6 & \textbf{88.6} & \textbf{2.1} & \textbf{1.5} \\\bottomrule[1.2pt]
    \end{tabular}
    \label{tab:comparison_appearance_based_methods}
\end{table}

\subsection{Ablation Study}
\label{sec:ablation}

\begin{table}[b]
    \footnotesize
    \setlength\tabcolsep{5pt}
    \centering
    \caption{Effectiveness and complexity of the proposed modules in terms of averaged rank-1 accuracy (\%) using the CASIA-B dataset.}
    \begin{tabular}{c|c|c|c|c|c|c}
        \toprule[1.2pt]
        \multirow{2}{*}{Model} & \multicolumn{4}{|c|}{Rank-1 Accuracy} & Params & FLOPs \\\cline{2-5}
        & NM & BG & CL & Avg & (M) & (G)\\\hline
        Baseline & 92.5 & 83.1 & 83.3 & 86.3 & 2.05 & 0.68 \\\hline
        Baseline w/JSFL & 93.9 & 87.2 & 87.2 & 89.5& 1.07 & 0.30 \\\hline
        Baseline w/VATL & 94.0 & 85.9 & 87.3 & 89.1 & 2.09 &0.72 \\\hline
        CAG (joint) & 94.9 & 87.8 & 88.1 & 90.3 & 1.17 & 0.38 \\\hline
        CAG (joint+bone) & \textbf{95.7} & \textbf{89.6} & \textbf{88.6} & \textbf{91.3} & 2.34 & 0.75 \\\bottomrule[1.2pt]
    \end{tabular}
    \label{tab:module effectiveness}
\end{table}

\noindent \textbf{Effectiveness of the proposed modules.} The effectiveness of the proposed modules is presented in \cref{tab:module effectiveness}. The baseline refers to replacing the CAG blocks with ordinary blocks in \cref{eq:graph conv} and using only the fixed topology for graph convolutions. The following main observations can be made: 1) Comparing the first three experiments, the proposed JSFL and VATL modules both contribute to recognition performance in all three conditions, which confirms the robustness and effectiveness of the dynamic filter learning and view-adaptive topology learning, respectively. 2) In the fourth experiment when applying JSFL and VATL modules together, not only further improves the recognition performance, but also requires fewer parameters and FLOPs than the baseline, which demonstrates the mutual promotion of each module and the efficiency of the proposed design. 3) Combining the joint and bone streams, the best accuracy is achieved, which confirms the complementary properties provided by the joint and bone streams.

\begin{table}[t]
    \centering
    \scriptsize
    \caption{Effectiveness of the proposed modules in terms of averaged rank-1 accuracy (\%) using the OU-MVLP dataset.}
    \begin{tabular}{c|c|c|c|c|c}
        \toprule[1.2pt]
        \multirow{2}{*}{Model} & \multirow{2}{*}{Baseline} & Baseline & Baseline & CAG & CAG \\ 
        & & w/JSFL & w/VATL & (joint) & (joint+bone) \\\hline 
        Rank-1 Accuracy (\%) & 26.2 & 39.2 & 38.7 & 45.2 & \textbf{56.4} 
        \\\bottomrule[1.2pt]
    \end{tabular}
    \label{tab:ablation_oumvlp}
\end{table}

Ablation experiments were also conducted using the OU-MVLP dataset (\cref{tab:ablation_oumvlp}), where each module still works well in a large-scale dataset.

\begin{table}[h]
    \centering
    \caption{JSFL impact on the CASIA-B dataset in terms of averaged rank-1 accuracy (\%).}
    \begin{tabular}{c|c|c|c|c}
        \toprule[1.2pt]
        \multirow{2}{*}{Model} & \multicolumn{4}{c}{Rank-1 Accuracy} \\\cline{2-5}
        & NM & BG & CL & Avg \\\hline
        Baseline & 92.5 & 83.1 & 83.3 & 86.3 \\\hline
        Baseline w/JSFL & 93.9 & 87.2 & 87.5 & 89.5 \\\hline
        Baseline w/JSFL & \multirow{2}{*}{91.6} & \multirow{2}{*}{75.5} & \multirow{2}{*}{78.6} & \multirow{2}{*}{81.9} \\
        (non-adaptive) & & & & \\\hline
        Baseline w/JSFL & \multirow{2}{*}{88.0} & \multirow{2}{*}{76.8} & \multirow{2}{*}{77.6} & \multirow{2}{*}{80.8} \\
        (global) & & & & \\
        \bottomrule[1.2pt]
    \end{tabular}
    \label{tab: JSFL}
\end{table}

\noindent \textbf{Investigation of JSFL impact.}
The JSFL impact was investigated by conducting experiments on the CASIA-B dataset to study the effects of adaptive filter learning and joint-level feature exploration individually. As shown in \cref{tab: JSFL}, the second experiment uses the proposed JSFL, the third experiment generates non-adaptive joint-specific filters, and the fourth experiment is equipped with adaptive filters learned at the global level. Therefore, JSFL (non-adaptive) refers to applying non-shared depth-wise convolutions to different joints, which is joint-specific but non-adaptive. JSFL (global) refers to learning only one group of weights for all joints, which is adaptive but not joint-specific. An interesting finding is that by taking the fine-grained or adaptive mechanism only, the performance is degraded, which indicates that by applying the joint-specific mining or adaptive learning mechanism only, both high model capacity and computational
efficiency is hardly achieved unless used together such as in JSFL.

\begin{table}[h]
    \centering
    \caption{Effectiveness of VATL topologies $G_1$, $G_2$ and $G_3$ on the CASIA-B dataset in terms of averaged rank-1 accuracy (\%).}
    \begin{tabular}{c|c|c|c|c|c|c}
        \toprule[1.2pt]
        \multicolumn{3}{c|}{Topology} & \multicolumn{4}{c}{Rank-1 Accuracy} \\\hline
        $G_1$ & $G_2$ & $G_3$ & NM & BG & CL & Avg \\\hline
        \checkmark& & & 93.8 & 85.7 & \textbf{87.3} & 88.8 \\\hline
        & \checkmark & & 93.7 & 84.7 & 87.0 & 88.5 \\\hline
        &&\checkmark& 92.5 & 83.1 & 83.3 & 86.3 \\\hline
        \checkmark&\checkmark & & 93.9 & 85.3 & 87.8 & 89.0 \\\hline
        \checkmark&&\checkmark& 93.8 & 85.8 & \textbf{87.3} &  89.0\\\hline
        &\checkmark&\checkmark& 93.8 & 84.7 & 87.1 & 88.5 \\\hline
        \checkmark&\checkmark&\checkmark& \textbf{94.0} & \textbf{85.9} & \textbf{87.3} & \textbf{89.1} \\\bottomrule[1.2pt]
    \end{tabular}
    \label{tab:VATL}
\end{table}

\noindent \textbf{Impact of VATL topologies.} The effectiveness of VATL topologies was investigated
by conducting ablation experiments using the CASIA-B dataset. In \cref{tab:VATL}, the following main observations can be made: 1) When applying only one topology, the adaptive learned $G_1$ and $G_2$ achieve better performance than the fixed $G_3$, which confirms the superiority of the proposed view-adaptive learning. 2) By combing them, better performance is achieved, which demonstrates that the topologies were designed at complementary levels. Thus, using all three topologies, the best performance can be achieved.

\begin{table}[b]
    \footnotesize
    \centering
    \setlength\tabcolsep{4pt}
    \caption{Performance of CAG combined with appearance-based methods in terms of averaged rank-1 accuracy (\%) using the CASIA-B dataset.}
    \begin{tabular}{c|c|c|c|c|c|c}
    \toprule[1.2pt]
    \multirow{2}{*}{Method} & Appearance & Ensemble & \multirow{2}{*}{NM} & \multirow{2}{*}{BG} &  \multirow{2}{*}{CL} & \multirow{2}{*}{Avg} \\
    & -based Method & Method & & & & \\\hline
    GaitSet \cite{gaitset} & - & - & 95.0 & 87.2 & 70.4 & 84.2 \\\hline
    GaitPart \cite{gaitpart} & - & - & 96.2 & 91.5 & 78.7 & 88.8 \\\hline
    GaitGL \cite{GaitGL} & - & - & 97.4 & 94.5 & 83.6 & 91.8 \\\midrule[1.2pt]
    GAITTAKE \cite{gaittake} & 3DCNN & Concat & 98.0 & 97.5 & 92.2 & 95.9 \\\hline
    BiFusion \cite{msgg} & GaitPart & FCs & 98.7 & 96.0 & 92.1 & 95.6 \\\hline
    Ours & 3D CNN & Concat & 98.6 & \textbf{97.9} & 93.0 & \textbf{96.5} \\\cline{1-7}
    Ours & GaitPart & FCs & \textbf{98.8} & 97.3 & \textbf{93.2} & 96.4 \\
    \bottomrule[1.2pt]
    \end{tabular}
    \label{tab:ensemble}
\end{table}

\noindent \textbf{Combining CAG with appearance-based methods.} The mutual promotion of CAG and appearance-based methods was investigated by integrating CAG with two appearance-based methods. In \cref{tab:ensemble}, for a fair comparison, we follow the ensemble settings as in GAITTAKE \cite{gaittake} and BiFusion \cite{msgg}. The results show that the combined network achieves better performances in all three scenarios, which demonstrates the complementary properties of motion modeling obtained from CAG and appearance learning obtained from appearance-based methods. And the best ensemble results also indicate that the proposed CAG is a better option to assist the appearance-based methods and make gait recognition more powerful.

\noindent \textbf{Comparison of a view-specific embedding method \cite{viewembedding1} and VATL.} As shown in Tab. \ref{tab:view_embedding}, we compare the proposed VATL module with a view-embedding method \cite{viewembedding1}, which aims to obtain transformed view-invariant features. We can see that the proposed view-adaptive topologies are more effective in dealing with view variations.

\begin{table}[h]
    \centering
    \caption{{Comparison of the proposed VATL module with a view-specific embedding method\cite{viewembedding1} using the CASIA-B dataset in terms of averaged rank-1 accuracy (\%).}}
    \begin{tabular}{c|c|c|c|c}
    \hline
        Method & NM & BG & CL & Avg \\\hline
        Baseline & 92.5 & 83.1 & 83.3 & 86.3 \\\hline
        Baseline w/view-specific embedding \cite{viewembedding1} & 93.0 & 83.2 & 83.3 & 86.5 \\\hline
        Baseline w/VATL & \textbf{94.0} & \textbf{87.8} & \textbf{87.3} & \textbf{89.1} \\\hline
    \end{tabular}
    \label{tab:view_embedding}
\end{table}

\subsection{HyperParameter Configurations}
\noindent \textbf{VATL Configurations.} The impact of different settings of coefficients $g_1$, $g_2$, and $g_3$ in VATL on the performance is presented in \cref{tab:VATL config}. It is observed that a balanced coefficient combination achieves better performance than an unbalanced one. Considering the first four experiments, the fourth model (90.3\%) outperforms the other three models when the sum of the adaptive topology coefficients ($g_1$ and $g_2$) equals the fixed topology coefficient ($g_3$). This result indicates the importance of balancing adaptive learning and general representation learning. The last two experiments also conform to the law that the sixth model (balanced) outperforms the fifth model (unbalanced). Finally, $g_1$, $g_2$ and $g_3$ were set to $\frac{1}{2}$, $\frac{1}{2}$, and 1, respectively, to achieve the best performance.

\begin{table}[h]
    \centering
    \caption{Impact of settings of coefficients $g_1$ ($G_1$), $g_2$ ($G_2$) and $g_3$ ($G_3$) in VATL on the performance in terms of averaged rank-1 accuracy (\%) using the CASIA-B dataset \cite{CAISA-B}.}
    \begin{tabular}{c|c|c|c|c|c|c}
    \toprule[1.2pt]
        $g_1$ & $g_2$ & $g_3$ & NM & BG & CL & Avg  \\\hline
        1 & 1 & 1 & 94.5 & 87.3 & 87.3 & 89.7 \\\hline
        1 & 1 & $\frac{4}{5}$ & 94.3 & 87.1 &	87.2 & 89.5 \\\hline
        1 & 1 & $\frac{1}{2}$ & 94.2 & 87.2 & 87.1 & 89.5 \\\hline
        $\frac{1}{2}$ & $\frac{1}{2}$ & 1 & \textbf{94.9} & 87.8 & \textbf{88.1} & \textbf{90.3} \\\hline
        $\frac{1}{4}$ & $\frac{1}{4}$ & 1 & 94.3 & 87.0 & 87.6 & 89.6 \\\hline
        $\frac{1}{4}$ & $\frac{1}{4}$ & $\frac{1}{2}$ & 94.4 & \textbf{88.3} & 87.5 & 90.1 \\
    \bottomrule[1.2pt]
    \end{tabular}
    \label{tab:VATL config}
\end{table}

\begin{table}[t]
    \small
    \centering
    \caption{Impact of settings of $K_T$, $\alpha$ and $r$ in JSFL on the performance in terms of averaged rank-1 accuracy (\%) and complexity using the CASIA-B \cite{CAISA-B} dataset.}
    \begin{tabular}{c|c|c|c|c|c|c|c|c}
    \toprule[1.2pt]
        \multirow{2}{*}{$K_T$} & \multirow{2}{*}{$\alpha$} & \multirow{2}{*}{$r$} & \multirow{2}{*}{NM} & \multirow{2}{*}{BG} & \multirow{2}{*}{CL} & \multirow{2}{*}{Avg} & param & FLOPs  \\
        & & & & & & & (M) & (G) \\\hline
        3 & 2 & 8 & 94.1 & 87.0 & 87.4 & 89.5 & 1.14 & 0.37 \\\hline
        5 & 2 & 8 & 94.5 & 87.2 & 87.8 & 89.8 & 1.14 & 0.38 \\\hline
        9 & 2 & 8 & \textbf{94.9} & 87.8 & \textbf{88.1} & \textbf{90.3} & 1.14 & 0.38 \\\midrule[1.2pt]
        9 & 2 & 8 & \textbf{94.9} & 87.8 & \textbf{88.1} & \textbf{90.3} & 1.14 & 0.38 \\\hline
        9 & 4 & 8 & \textbf{94.9} & 87.6 & 88.0 & 90.2 & 1.15 & 0.39 \\\midrule[1.2pt]
        9 & 2 & 4 & 94.8 & \textbf{88.0} & \textbf{88.1} & \textbf{90.3} & 1.70 & 0.51 \\\hline
         9 & 2 & 8 & \textbf{94.9} & 87.8 & \textbf{88.1} & \textbf{90.3} & 1.14 & 0.38 \\\hline
        9 & 2 & 16 & 94.2 & 86.7 & 87.4 & 89.4 & 0.86 & 0.31 \\
    \bottomrule[1.2pt]
    \end{tabular}
    \label{tab:JSFL}
\end{table}

\noindent \textbf{JSFL Configurations.} The impact of different settings of temporal kernel size $K_T$, temporal inflation ratio $\alpha$, and channel reduction ratio $r$ in JSFL on the
performance is presented in \cref{tab:JSFL}.The following observations can be made: 1) By comparing the first three experiments ($K_T=3$, 5, and 9, respectively), it is observed that the performance increases as the value of $K_T$ increases, which indicates that large temporal receptive fields benefit from modeling rich temporal clues in gait modeling. Thus, $K_T$ was set to 9 to achieve the best performance. 2) By comparing the fourth ($\alpha=2$) and the fifth ($\alpha=4$) experiments, it is observed that the proposed model with $\alpha=2$ achieves better performance and lower complexity than that with $\alpha=4$. Thus, $\alpha=2$ was set in the proposed network. 3) By comparing the last three experiments ($r=4$, 8, and 16), it is observed that the proposed model with $r=4$ achieves the best average accuracy but it is much more complex than those with $r=8$ and $r=16$. The model with $r=16$ exhibits the worst performance because a large channel reduction ratio degrades the representation ability. Finally, considering both the performance and complexity, we set $r=8$ in our network.

\begin{table}[h]
    \centering
    \caption{Impact of the loss function settings $\lambda_1$ ($L_{tri}$), $\lambda_2$ ($L_{circle}$) and $\lambda_3$ ($L_{CE}^{view}$) on the performance in terms of averaged rank-1 accuracy (\%) using the CASIA-B \cite{CAISA-B} dataset.}
    \begin{tabular}{c|c|c|c|c|c|c}
    \toprule[1.2pt]
        $\lambda_1$ & $\lambda_2$ & $\lambda_3$ & NM & BG & CL & Avg  \\\hline
        1 & 0 & 0 & 92.4 & 86.5 & 86.1 & 88.3 \\\hline
        1 & 0 & 1 & 92.9 & 87.7 & 87.3 & 89.3 \\\hline
        0 & 1 & 1 & 91.5 & 84.2 & 84.7 & 86.8 \\\hline
        1 & 1 & 1 & 93.8 & 87.6 & 87.3 & 89.5 \\\hline
        1 & 1 & 0.1 & 94.1 & 87.6 & 87.6 & 89.8 \\\hline
        0.5 & 0.5 & 0.1 & 94.0 & 87.5 & 87.5 & 89.7 \\\hline
        0.3 & 0.7 & 0.1 & 93.7 & 86.4 & 86.7 & 88.9 \\\hline
        0.7 & 0.3 & 0.1 & 94.5 & 87.6 & 87.9 & 90.0 \\\hline
        0.9 & 0.1 & 0.1 & \textbf{94.9} & \textbf{87.8} & \textbf{88.1} & \textbf{90.3} \\
    \bottomrule[1.2pt]
    \end{tabular}
    \label{tab:loss config}
\end{table}

\noindent \textbf{Weights in the loss function.} The impact of weights used in the loss function on the performance is presented in \cref{tab:loss config}. The following observations can be made: {1) Comparing the results in the first and second rows, based on the triplet loss, using the view-classification loss could further improve performance.} 2) Comparing the results in the second and third rows, based on the view-classification loss, using triplet loss achieves better performance than using circle loss; when both are used, better performance can be achieved as shown in the fourth row. 3) By increasing the weight of triplet loss ($\lambda_1$) and decreasing the weight of circle loss ($\lambda_2$) simultaneously, the performances can be further improved. 4) By decreasing the weight of cross entropy loss ($\lambda_3$) used in view classification, the recognition performance can be improved, which indicates that decreasing $\lambda_3$ helps the model focus on modeling identity-related features. 

\begin{figure}[t]
	\centering
	\subfigure[A sequence obtained from subject '100'.]{
		\begin{minipage}[t]{0.5\linewidth}
			\centering
			\includegraphics[width=1.8in]{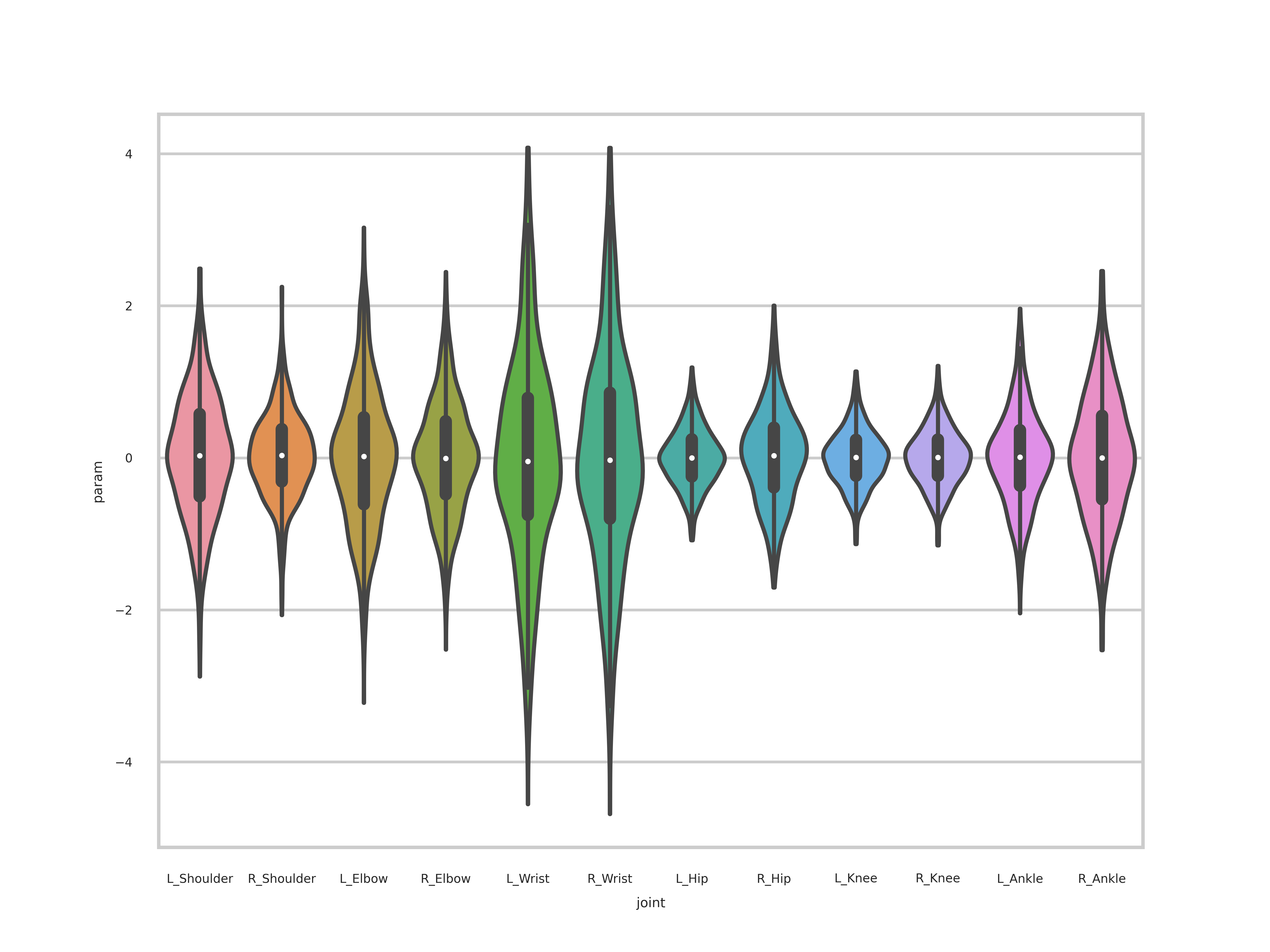}
       \label{fig:100_1}
		\end{minipage}
	}%
	\subfigure[A different sequence obtained from subject '100'.]{
		\begin{minipage}[t]{0.45\linewidth}
			\centering
			\includegraphics[width=1.8in]{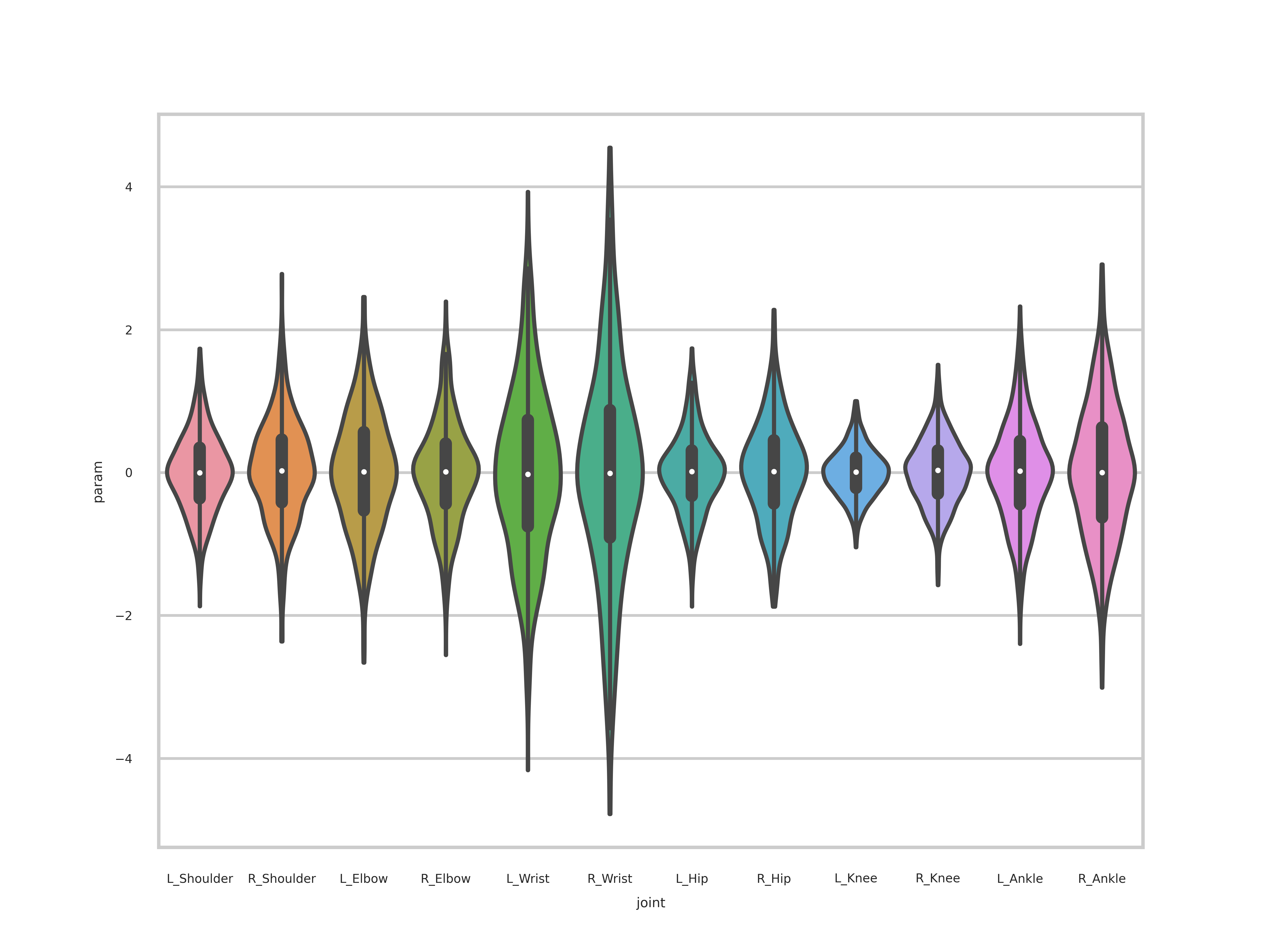}
           \label{fig:100_2}
		\end{minipage}
	}%
 
	\subfigure[Sequence obtained from subject '102'.]{
		\begin{minipage}[t]{0.5\linewidth}
			\centering
			\includegraphics[width=1.8in]{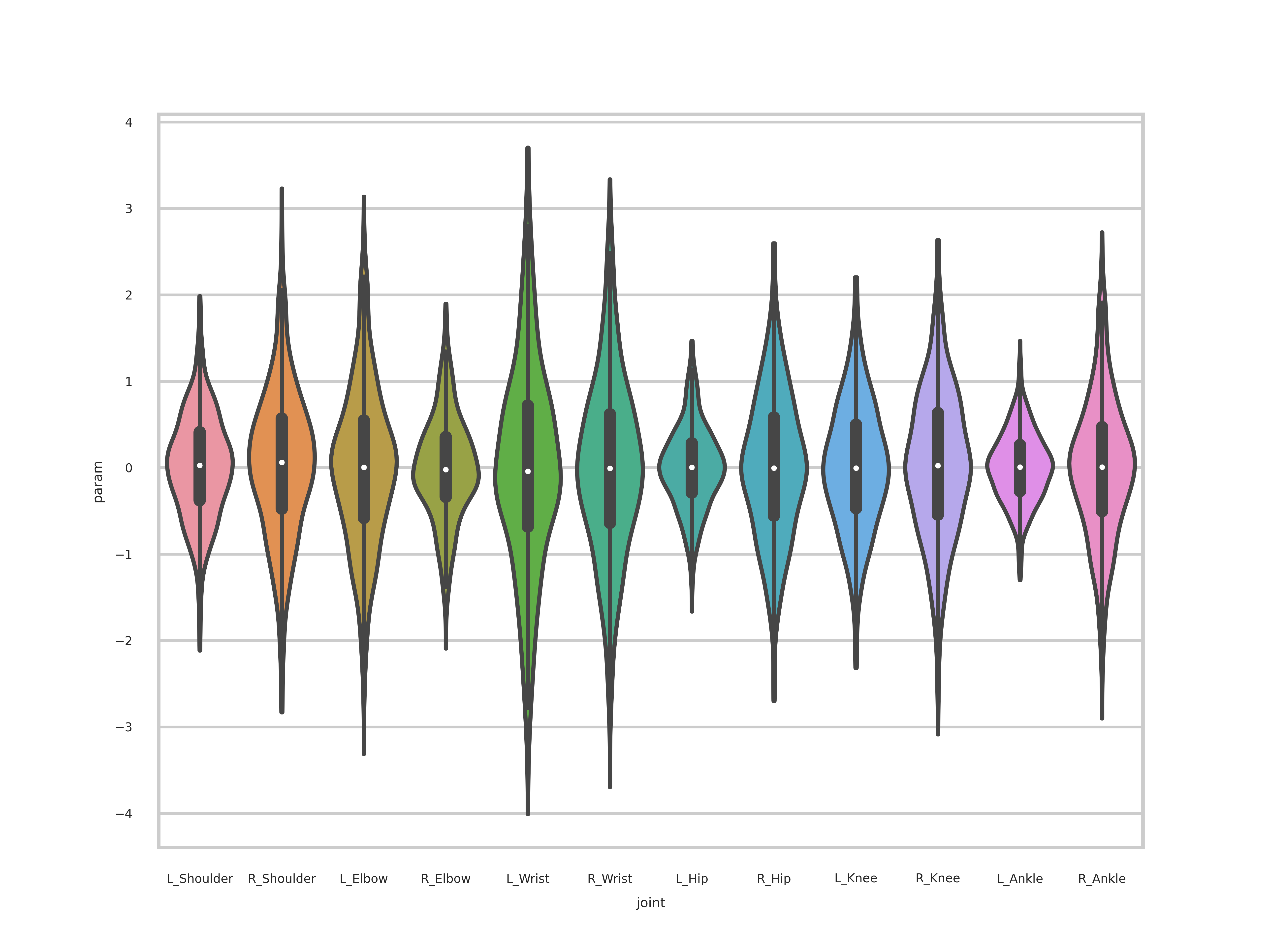}
       \label{fig:102}
		\end{minipage}
	}%
	\subfigure[Sequence obtained from subject '103'.]{
		\begin{minipage}[t]{0.45\linewidth}
			\centering
			\includegraphics[width=1.8in]{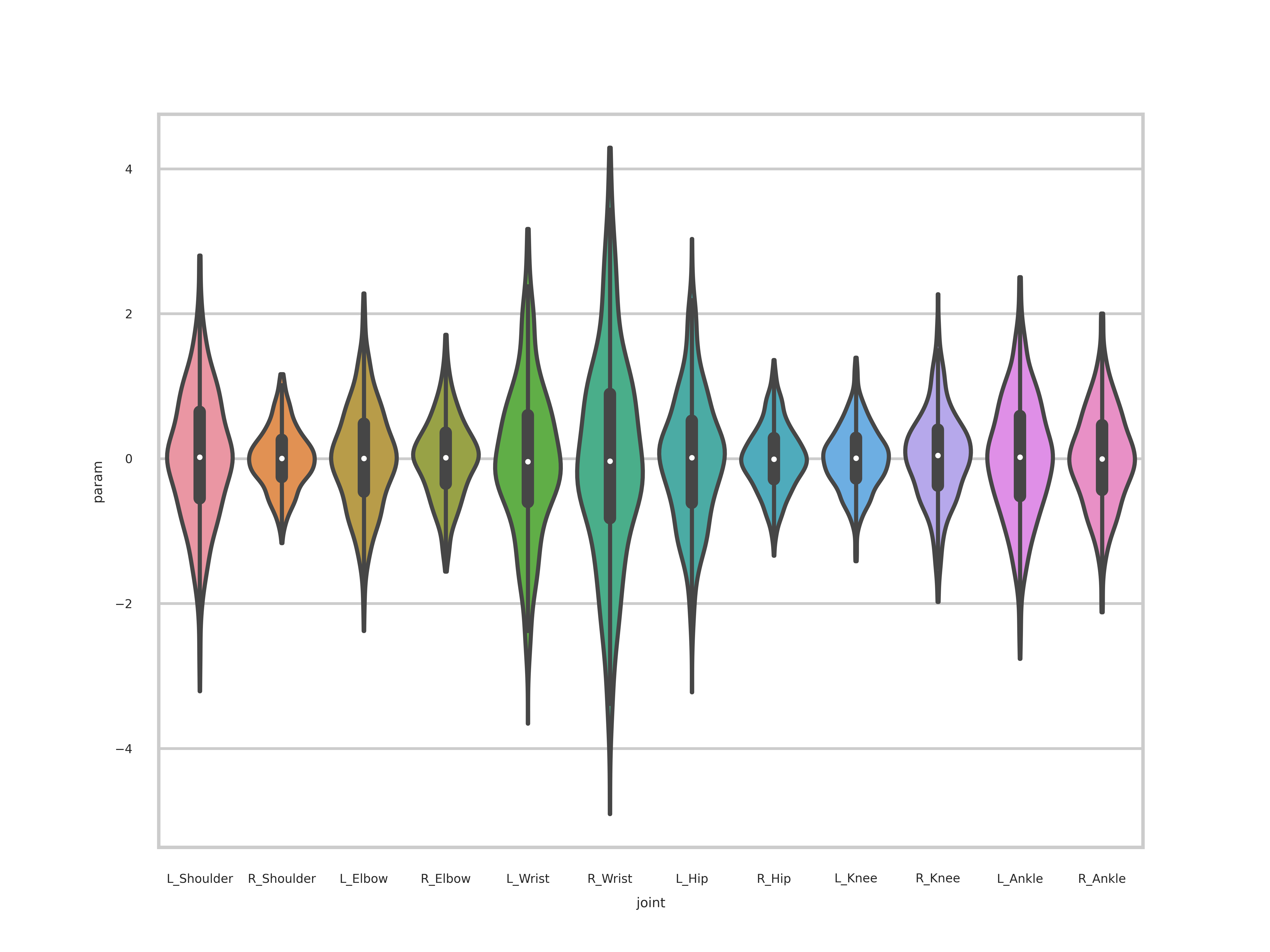}
           \label{fig:103}
		\end{minipage}
	}%
	\centering
        \caption{Visualized statistics of the learned filters in the CASIA-B dataset using a violin plot, which plots the parameter distribution of body joints on the arms and legs in the last CAG block. The area of the violin plot denotes the data range, and the bandwidth represents the probability density of data for different values. In each plot, the black box and the range of the slim black line represent the interquartile range and the 95\% confidence interval respectively. Best viewed with zoom in.}
        \label{fig:parameter distribution}
\end{figure}

\subsection{Visualization}
\noindent \textbf{Learned filters.} The distribution of the learned filters in JSFL can be visualized using a violin plot, as shown in \cref{fig:parameter distribution}. The following observations can be made: 1) {\cref{fig:100_1} and \cref{fig:100_2} represent the data distribution of the learned filters from the same person but different sequences, i.e., the two sequences are captured under different camera viewpoints and walking conditions. Therefore, there exist intra-subject variations in the two sequences. To adapt to the variations, the learned filters for the two sequences have slight differences.} 2) By comparing the learned filters of two different subjects '102' and '103' (\cref{fig:102} and \cref{fig:103}), it is observed that the filters are learned diversely, which is mainly due to the personalized walking styles of different subjects. 

With the help of visualization, we can realize the adaptation ability of JSFL to cope with customized characteristics in different sequences, which offers a possible scheme to model complex gait patterns.

\begin{figure}[t]
    \centering
    \includegraphics[width=\linewidth]{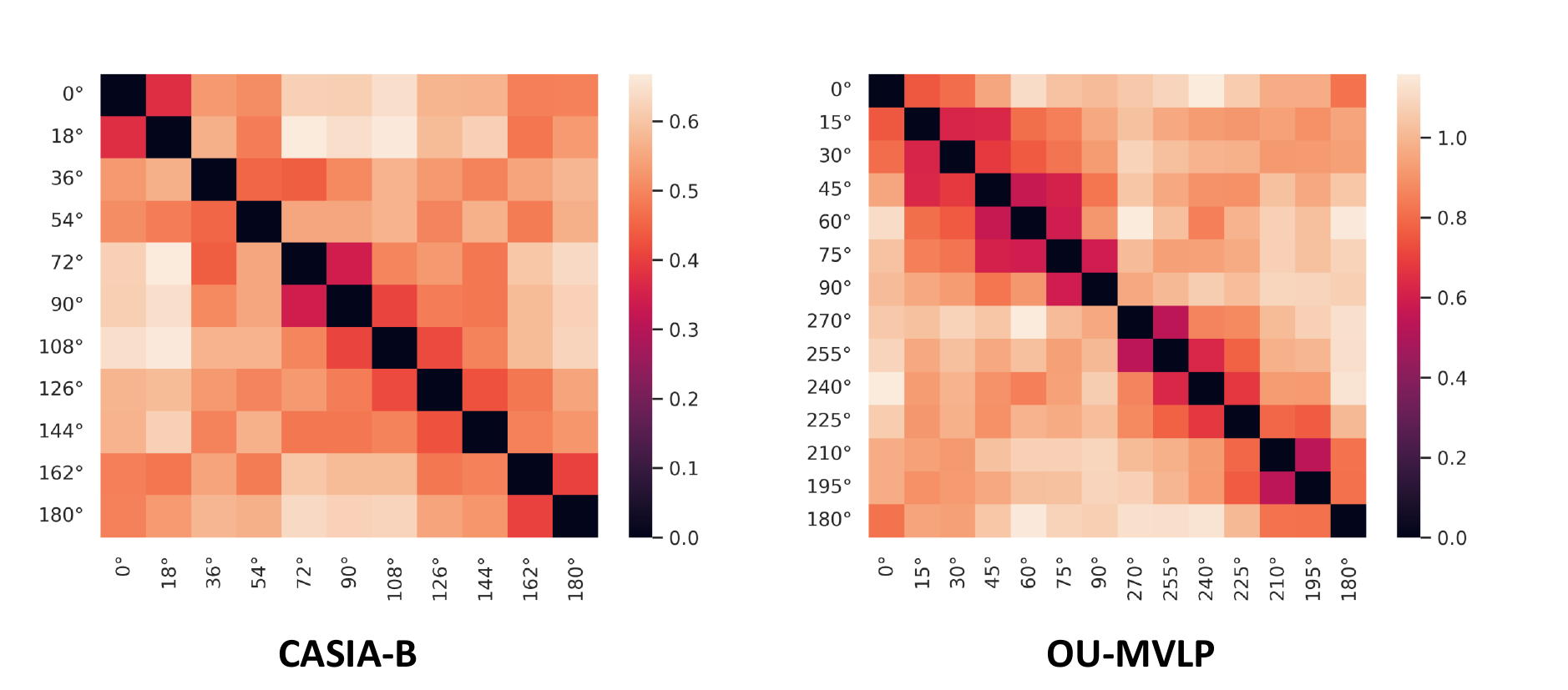}
    \caption{Topology correlations in $G_{set}$ of the joint stream in CASIA-B and OU-MVLP datasets. The correlation is measured by the mean square error. Darker color indicates a stronger correlation. Best viewed in color.}
    \label{fig:view relation}
\end{figure}

\noindent \textbf{View-adaptive topologies.} In \cref{fig:view relation}, the topology correlations in view-related topology set $G_{set}$ in the CASIA-B and OU-MVLP datasets are visualized. The following two interesting findings that adhere to human intuitions are summarized as 1) The main diagonal line of the two heatmaps indicates that the view-adjacent topologies generally exhibit stronger correlations than the view-distant topologies, indicating that sequences in adjacent views possess similar spatial-temporal characteristics. 2) The anti-diagonal lines of the two heatmaps indicate that the mirror-view topologies exhibit relatively strong correlations (e.g., $0^\circ$ and $180^\circ$, $18^\circ$ and $162^\circ$ in the CAISA-B dataset; $0^\circ$ and $180^\circ$, $60^\circ$ and $240^\circ$ in the OU-MVLP dataset), indicating that sequences in mirror views have related features. Consequently, the high response regions on the heatmaps resemble an 'X' format.

\section{Conclusion}
In this paper, a CAG convolutional network for skeleton-based gait recognition was proposed. For each sequence, CAG automatically produces dynamic joint-specific filters to describe personalized fine-grained walking features and learns a view-adaptive topology to fit customized properties under various view conditions. In this way, CAG achieves great adaptation ability in complex scenarios, using the two most popular datasets (i.e., CASIA-B and OU-MVLP), demonstrating its superiority over previous methods. Furthermore, the potential of integrating skeleton-based and appearance-based methods was investigated. Further investigation on combining these two methods will be conducted in future work.

\section*{Acknowledgement}
This research was supported by the NSFC (grant no.61773176).

{\small
\bibliographystyle{IEEEtran}
\bibliography{rf}
}

\begin{IEEEbiography}[{\includegraphics[width=1in,height=1.25in,clip,keepaspectratio]{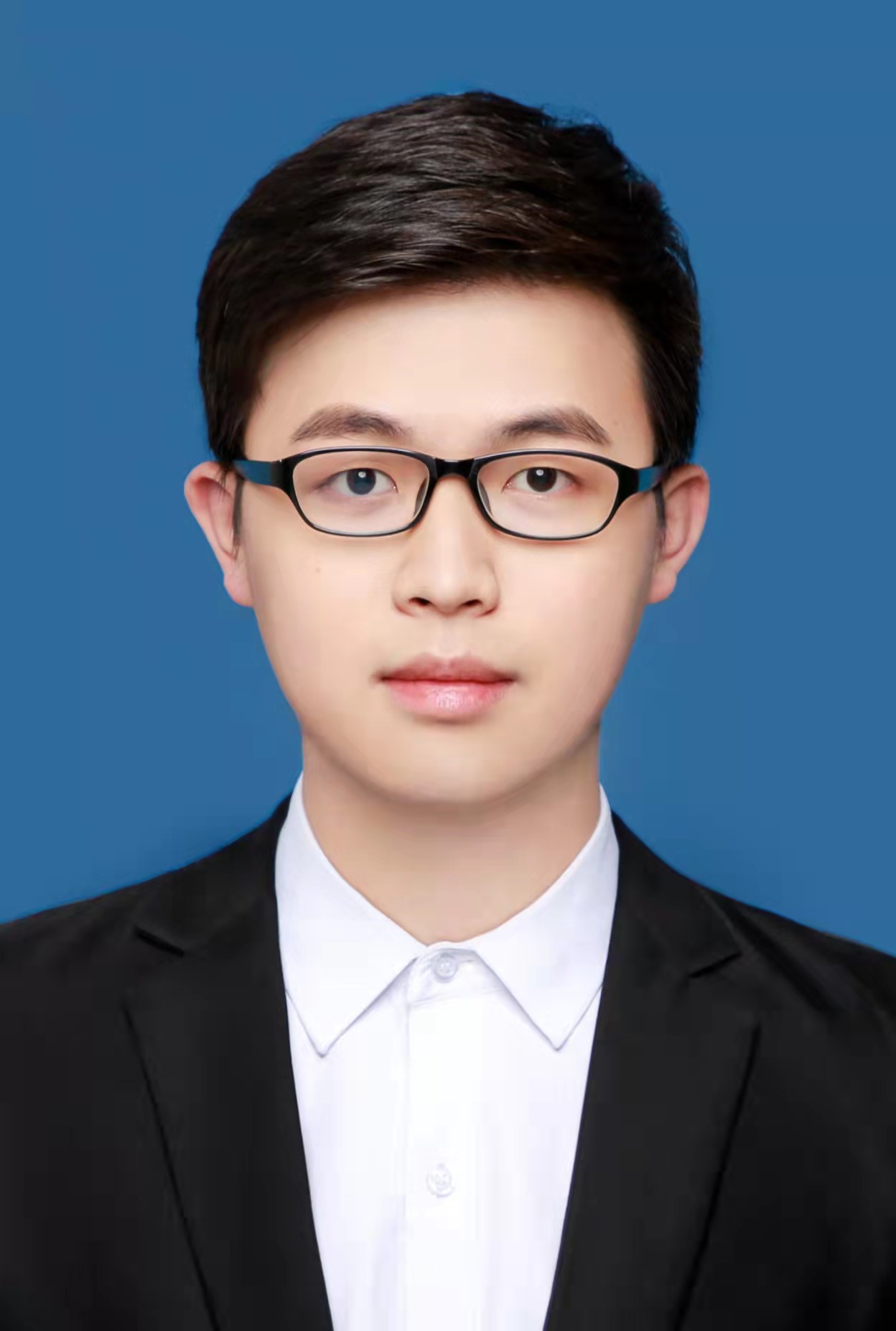}}] {Xiaohu Huang} received the B.E. and M.E. degree in School of Electronic Information and Communications from Huazhong University of Science and Technology (HUST), Wuhan, China, in 2020 and 2023. His current research areas include computer vision and machine learning.
\end{IEEEbiography}

\begin{IEEEbiography}[{\includegraphics[width=1in,height=1.25in,clip,keepaspectratio]{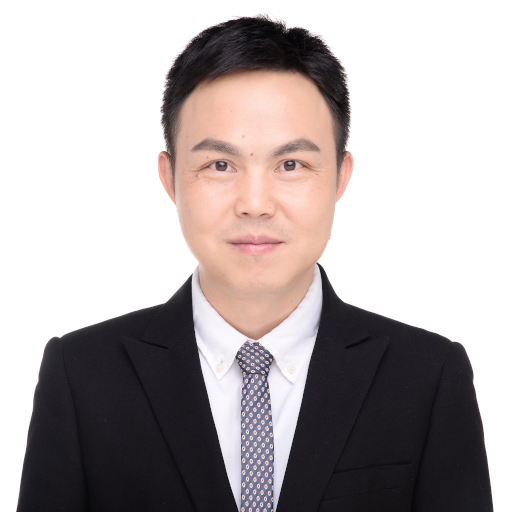}}] {Xinggang Wang (M’17)} received the B.S. and Ph.D. degrees in Electronics and Information Engineering from Huazhong University of Science and Technology (HUST), Wuhan, China, in 2009 and 2014, respectively. He is currently an Associate Professor with the School of Electronic Information and Communications, HUST. His research interests include computer vision and machine learning. He services as associate editors for Pattern Recognition and Image and Vision Computing journals and an editorial board member of Electronics journal.
\end{IEEEbiography}

\begin{IEEEbiography}[{\includegraphics[width=1in,height=1.25in,clip,keepaspectratio]{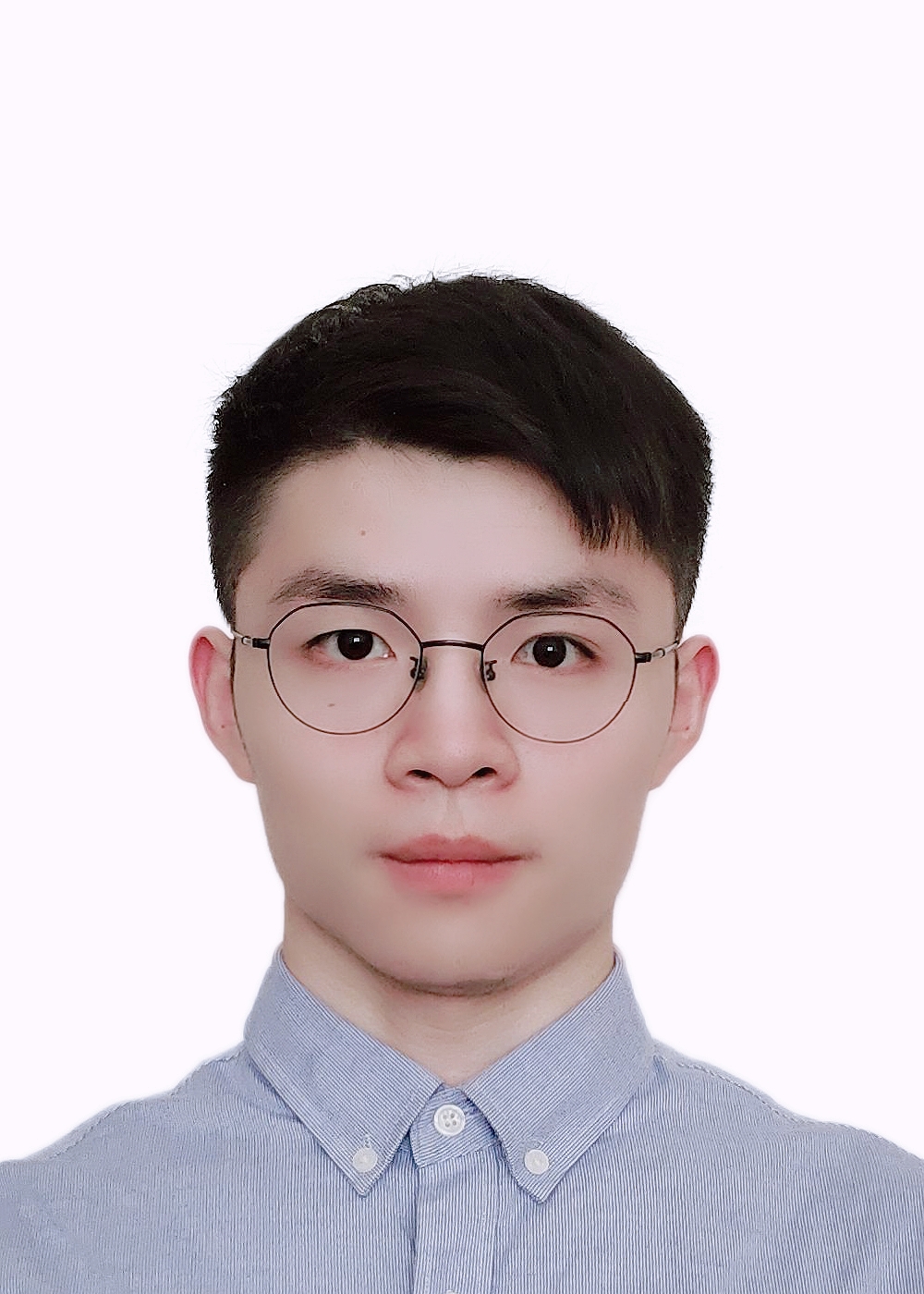}}] {Zhidianqiu Jin} received the M.E. degree in School of Electronic Information and Communications from Huazhong University of Science and Technology (HUST), Wuhan, China, in 2021. His current research areas include computer vision and machine learning.
\end{IEEEbiography}

\begin{IEEEbiography}[{\includegraphics[width=1in,height=1.25in,clip,keepaspectratio]{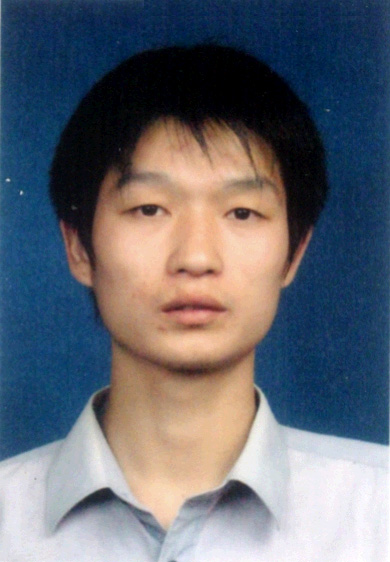}}]{Bo Yang} received the Master degree in School of mathematics and statistics form Wuhan University,Wuhan, China. He is currently the senior engineer of Wuhan FiberHome Digital Technology Co., Ltd. His research interests include computer vision and data mining.
\end{IEEEbiography}

\begin{IEEEbiography}[{\includegraphics[width=1in,height=1.25in,clip,keepaspectratio]{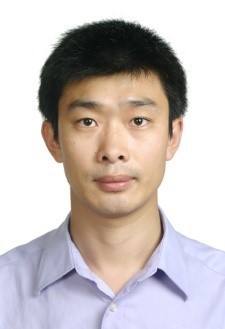}}]{Botao He} received the Ph.D. degree in School of Optical and Electronic Information from Huazhong University of Science and Technology (HUST), Wuhan, China. He is currently the deputy general manager of Wuhan FiberHome Digital Technology Co., Ltd. His research interests include computer vision and data mining.
\end{IEEEbiography}

\begin{IEEEbiography}[{\includegraphics[width=1in,height=1.25in,clip,keepaspectratio]{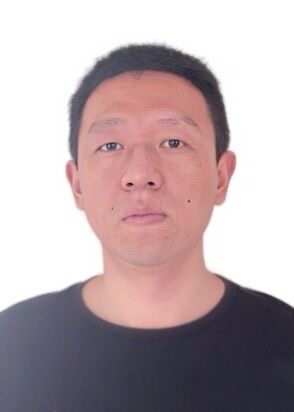}}] {Bin Feng} received the B.S. and Ph.D. degrees in School of Electronics and Information Engineering from Huazhong University of Science and Technology (HUST), Wuhan, China, in 2001 and 2006, respectively. He is currently an Associate Professor with the School of Electronic Information and Communications, HUST. His research interests include computer vision and intelligent video analysis.
\end{IEEEbiography}

\begin{IEEEbiography}[{\includegraphics[width=1in,height=1.25in,clip,keepaspectratio]{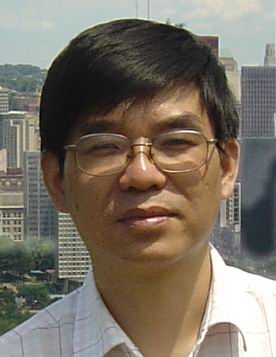}}] {Wenyu Liu (SM'15)} received the B.S. degree in Computer Science from Tsinghua University, Beijing, China, in 1986, and the M.S. and Ph.D. degrees, both in Electronics and Information Engineering, from Huazhong University of Science and Technology (HUST), Wuhan, China, in 1991 and 2001, respectively. He is now a professor and associate dean of the School of Electronic Information and Communications, HUST. His current research areas include computer vision, multimedia, and machine learning.
\end{IEEEbiography}

\end{document}